\documentclass[a4paper,fleqn]{cas-sc}

\usepackage[authoryear]{natbib}
\usepackage{float}
\usepackage{booktabs}
\usepackage{tabularx}
\usepackage{array}
\usepackage{multirow}
\usepackage{pifont}
\usepackage{graphicx}

\def\tsc#1{\csdef{#1}{\textsc{\lowercase{#1}}\xspace}}
\tsc{WGM}
\tsc{QE}

\begin{document}
\let\WriteBookmarks\relax
\def\floatpagepagefraction{1}
\def\textpagefraction{.001}

\shorttitle{}    

\shortauthors{}  

\title [mode = title]{Brain-Atlas-Guided Generative Counterfactual Attention for Explainable Cognitive Decline Diagnosis Using Multimodal Connectomes}  

\tnotemark[1] 

\tnotetext[1]{} 

%

\author[1]{Xiongri Shen}
\author[1]{Jiaqi Wang}
\author[2]{Zhenxi Song}
\author[1]{Yi Zhong}
\author[1]{Leilei Zhao}
\author[1]{Xin He}
\author[3]{Baiying Lei}
\author[2]{Zhiguo  Zhang}
\cortext[cor1]{Corresponding author: 
zhiguozhang@hit.edu.cn}
\cormark[1]





\fntext[1]{Code and models are available at: \href{https://github.com/SXR3015/GCAN}{GCAN}}

\address[1]{Department of Computer Science and Technology, Harbin Institute of Technology (Shenzhen), Shenzhen, 518055, China}
\address[2]{School of Intelligence Science and Engineering, College of Artificial Intelligence, Harbin Institute of Technology, Shenzhen, 518055, China}
\address[3]{School of Biomedical Engineering, National-Regional Key Technology Engineering Laboratory for Medical Ultrasound, Guangdong Key Laboratory for Biomedical, Measurements and Ultrasound Imaging, Shenzhen University Medical School, Shenzhen University, Shenzhen, 518055, China}









\begin{abstract}
Mild cognitive impairment (MCI) and subjective cognitive decline (SCD) are closely associated with the early Alzheimer's disease continuum, where accurate and explainable diagnosis is important for early risk assessment and intervention. Existing connectome-based deep learning models can improve classification performance but often provide limited insight into disease-related functional and structural connectivity changes. This paper proposes an atlas-knowledge-guided Generative Counterfactual Attention-guided Network (GCAN) for explainable cognitive decline diagnosis using multimodal brain connectomes. GCAN formulates diagnosis as a source-to-target counterfactual generation problem, where target-label connectomes are generated from source-label inputs and their differences are used to construct counterfactual attention maps. To preserve connectome topology, an Atlas-aware Bidirectional Transformer (AABT) performs network-level token encoding and decoding under brain-atlas constraints. The framework is further extended from functional connectivity (FC) to joint functional and structural connectivity (SC) modeling, enabling counterfactual analysis of complementary functional reorganization and structural topology changes. Experiments on hospital-collected and ADNI datasets show that GCAN achieves competitive performance across HC vs. SCD, HC vs. MCI, and SCD vs. MCI classification tasks. Visualization, circular connectome analysis, CAM-based comparison, ablation studies, and confidence interval analysis further support the interpretability and reliability of the proposed framework. Modality-specific FC and SC pre-trained classifiers are used to provide target-state priors for counterfactual generation while being separated from the downstream diagnostic classifier to prevent data leakage.
\end{abstract}


\begin{highlights}
\item A generative counterfactual framework explains cognitive decline diagnosis.
\item Atlas-aware bidirectional Transformers model structured brain connectomes.
\item Counterfactual attention localizes transition-related FC and SC patterns.
\item Multimodal FC--SC modeling captures complementary connectome information.
\item Experiments on hospital and ADNI cohorts support competitive performance.
\end{highlights}

\begin{keywords}
 Atlas-knowledge-guided learning \sep  Explainable artificial intelligence \sep Counterfactual reasoning  \sep Cognitive decline diagnosis \sep Multimodal brain connectivity  \sep Generative attention-guided network
\end{keywords}

\maketitle


\section{Introduction}
\label{sec:introduction}

Alzheimer's disease (AD) is a progressive neurodegenerative disorder characterized by cognitive and functional decline. Mild cognitive impairment (MCI) and subjective cognitive decline (SCD) are widely regarded as early clinical stages or risk states preceding AD \cite{petersen1999mild,jessen2014conceptual}, and their accurate identification is therefore important for early screening, risk assessment, and timely intervention. Resting-state functional magnetic resonance imaging (rs-fMRI) provides a non-invasive way to characterize spontaneous neural activity and functional organization \cite{fox2010clinical}. In particular, functional connectivity (FC), commonly estimated by calculating the correlations among regional blood-oxygen-level-dependent time series, has been extensively used to reveal abnormal interactions among brain regions in SCD and MCI \cite{ramirez2021functional,liebe2022investigation}. Diffusion MRI (dMRI), on the other hand, characterizes white-matter microstructure and enables the construction of structural connectivity (SC). Together, FC and SC provide complementary views of brain network alterations and have been widely studied in complex brain network analysis \cite{bullmore2009complex}. FC reflects functional synchronization and dynamic network reorganization, whereas SC describes the anatomical pathways that constrain information communication across regions.

Deep learning has substantially advanced automatic diagnosis of cognitive decline based on FC or other neuroimaging-derived features. Convolutional neural networks, residual networks, graph neural networks, and Transformer-based models have been introduced to extract discriminative representations from brain connectivity matrices \cite{li2021virtual,zuo2024u,xia2025interpretable,han2026rethinking}. Although these methods often improve classification performance, most of them remain black-box predictors. They usually provide limited information about which brain connections support a decision, how a sample may move from one cognitive state to another, and whether the highlighted connections are consistent with known neurodegeneration-related brain networks. This lack of interpretability is particularly problematic in clinical neuroimaging, where recent multimodal dementia studies have also emphasized the need for transparent decision-making mechanisms \cite{park2025explainable,han2026rethinking}.

Existing explanation methods provide only partial solutions. Recent reviews of XAI in AD neuroimaging have summarized commonly used post-hoc explanation techniques, such as SHAP, LIME, Grad-CAM, and layer-wise relevance propagation, and emphasized that insufficient interpretability remains a key barrier to clinical translation \cite{khosroshahi2025explainable}.
Gradient-based and activation-based visualization methods, such as Grad-CAM and Score-CAM, have been widely used to highlight regions contributing to classification outputs \cite{selvaraju2017grad,wang2020score}. Attention mechanisms have also been used to reweight informative features and improve model performance. However, these approaches generally produce post-hoc explanations conditioned on predicted class labels. When a classifier is uncertain or incorrect, the generated explanation can be dominated by the prediction result itself rather than by the intrinsic difference between cognitive states. Moreover, conventional attention maps typically describe feature importance in a correlational manner and do not explicitly model the transition between a source state and a target state. For early cognitive decline, where the differences between HC, SCD, and MCI can be subtle, an explanation should ideally capture the minimal and disease-relevant changes required to transform one connectome pattern into another.

Counterfactual reasoning offers a promising paradigm for interpretable neuroimaging analysis by asking what minimal changes would alter a model decision \cite{wachter2018counterfactual}. Instead of only asking which features support the current prediction, counterfactual reasoning asks how an input would need to change in order to be recognized as another target class. Such a formulation can reveal decision-relevant changes near the boundary between cognitive states and can naturally represent the transition from HC to SCD or MCI. Counterfactual explanation has been explored in structural MRI and lesion analysis \cite{oh2022learn,ren2023punctate}, and recent studies have further introduced counterfactual explanations into SC--FC coupling analysis for brain disorder prediction \cite{huang2025local}. However, its application to brain connectivity remains challenging. FC and SC matrices are not ordinary images; they are highly structured connectome representations constrained by brain atlases, network partitions, and symmetric region-to-region relationships. A counterfactual generator for connectomes must therefore preserve both local connection patterns and global network topology while producing target-class-specific changes.

In our MICCAI 2024 conference paper, we introduced a Generative Counterfactual Attention-guided Network (GCAN) to generate target-label FC from source-label FC and define counterfactual attention as the difference between the generated target FC and the original source FC. We further designed an Atlas-aware Bidirectional Transformer (AABT) to encode and decode FC according to atlas-defined brain networks, thereby improving the generation of structured FC matrices. For the journal extension, we substantially expand the original framework in three aspects. First, we reorganize GCAN into a more general counterfactual connectome reasoning framework that explicitly formulates source-to-target generation, attention aggregation, and attention-guided diagnosis. Second, we extend the model from single-modal FC explanation to multimodal structure--function counterfactual reasoning by jointly modeling FC and SC. Third, we provide more comprehensive experiments, including diagnostic comparisons, counterfactual attention visualization, circular connectome analysis, CAM-based interpretation comparison, matrix- and edge-level synthesis quality evaluation, fold-level confidence interval analysis, and ablation studies in both single-modal and multimodal settings.

The main contributions of this journal version are summarized as follows:
\begin{itemize}
    \item We propose a generative counterfactual attention framework for explainable cognitive decline diagnosis. By generating target-label connectomes from source-label inputs and subtracting the source connectomes, GCAN explicitly identifies disease-relevant connection changes associated with HC, SCD, and MCI transitions.
    \item We design an Atlas-aware Bidirectional Transformer (AABT) for structured connectome generation. AABT performs network-wise tokenization and inverse token decoding according to brain atlas partitions, allowing the model to preserve connectome topology while learning long-range region-to-region dependencies.
    \item  We extend counterfactual reasoning from single-modal FC to multimodal FC--SC joint modeling. The proposed multimodal GCAN learns modality-specific counterfactual attention for both FC and SC, enabling complementary interpretation of functional reorganization and structural topology changes during cognitive decline.
    \item We establish a multi-stage training strategy with modality-specific FC and SC pre-trained classifiers. Auxiliary datasets, including SLIM for FC pre-training and BJE for SC pre-training, are used to learn diagnostic priors for target-label connectome generation. The pre-trained classifiers are separated from the downstream diagnostic classifier to reduce the risk of data leakage. Experiments in both single-modal and multimodal settings show that GCAN achieves competitive diagnostic performance and identifies biologically plausible disease-related networks.
    \item We provide additional interpretability and uncertainty analyses. The proposed counterfactual attention is compared with Grad-CAM and Score-CAM through circular connectome visualization, and fold-level 95\% confidence intervals are reported to describe cross-validation uncertainty. These analyses provide a more cautious and transparent evaluation of the proposed framework.
\end{itemize}

\section{Related Work}
\label{sec:related_work}

\subsection{Connectome-based diagnosis of cognitive decline}

Brain connectivity provides an important network-level representation for studying cognitive decline. FC captures temporal synchronization among brain regions and has been used to identify abnormal interactions in MCI and SCD \cite{ramirez2021functional,liebe2022investigation}. SC, commonly derived from diffusion MRI tractography or diffusion-derived structural measures, reflects white-matter anatomical organization and provides complementary information about structural degeneration \cite{bullmore2009complex,sporns2011networks}. Existing studies have reported that cognitive decline is associated with altered connectivity in high-order cognitive networks, including the default mode network (DMN), fronto-parietal network (FPN), and cingulo-opercular network (CON) \cite{mah2021distinct,ghanbari2023accumulation}. These findings motivate the development of computational models that can learn discriminative connectome features for early diagnosis.

Deep learning models have been widely adopted for connectivity-based diagnosis. CNNs and ResNets can treat connectivity matrices as two-dimensional inputs, while Transformer-based models can capture long-range dependencies among regional connections \cite{li2021virtual,zuo2024u}. Graph neural networks further model brain regions as nodes and connections as edges, making them suitable for connectome representation learning \cite{xia2025interpretable,han2026rethinking}. Although these methods are effective for classification, they often focus on predictive performance rather than explanation. In clinical applications, it is not sufficient to know whether a subject is classified as HC, SCD, or MCI; it is also important to know which network alterations drive the prediction and whether these alterations align with known mechanisms of neurodegeneration \cite{rudin2019stop,khosroshahi2025explainable}.

\subsection{Explainable learning for neuroimaging}

Explainable artificial intelligence has become increasingly important in medical image analysis, especially for neuroimaging-based Alzheimer's disease diagnosis \cite{khosroshahi2025explainable}. Gradient-based methods, class activation mapping, and attention visualization have been applied to highlight task-relevant regions in neuroimaging data \cite{selvaraju2017grad,wang2020score}. In FC-based diagnosis, attention mechanisms can assign higher weights to informative connections or regions \cite{woo2018cbam}. However, most of these methods are label-conditioned explanations derived from the final classifier. As a result, explanations may be sensitive to prediction errors and may not directly reflect the transformation between two cognitive states.

Counterfactual explanation provides a more explicit way to interpret model decisions. A counterfactual sample answers the question: what minimal change would make a source sample resemble a target class \cite{wachter2018counterfactual}. For cognitive decline, this formulation is clinically meaningful because disease progression can be viewed as a gradual transition between cognitive states. By constructing target-label FC or FC--SC representations from source-label inputs, counterfactual reasoning can identify the connection changes most responsible for the transition. Counterfactual explanation has been explored in structural MRI and lesion analysis \cite{oh2022learn,ren2023punctate}, and recent studies have further introduced counterfactual explanations into structure--function coupling analysis for brain disorder prediction \cite{huang2025local}. Nevertheless, generating realistic counterfactual connectomes is difficult because connectivity matrices have atlas-aware, symmetric, and network-structured properties. This motivates the proposed AABT-based counterfactual generator.

\subsection{Structure--function multimodal modeling}

FC and SC represent different but related aspects of the brain connectome. Functional abnormalities may emerge early as altered synchronization among distributed regions, whereas structural degeneration may reflect white-matter damage or anatomical disconnection \cite{bullmore2009complex,sporns2011networks}. Multimodal modeling can therefore provide a more complete view of cognitive decline than either modality alone. Recent studies have shown that jointly modeling modality-specific information and structure--function interactions can improve neurological disease analysis \cite{xia2025interpretable,huang2025local}. However, FC and SC differ in their statistical properties, noise characteristics, and biological meanings. Simple feature concatenation may not effectively capture structure--function coupling, and post-hoc explanations from concatenated features may fail to distinguish functional and structural contributions.

The journal extension addresses this limitation by extending GCAN to FC--SC joint counterfactual reasoning. Instead of only generating a target-label FC, the multimodal framework generates target-label FC and SC simultaneously and constructs modality-specific counterfactual attention maps. The resulting attention maps allow the model to capture functional reorganization, structural topology changes, and their shared disease-related patterns.

\section{Methodology}
\label{sec:method}

\subsection{Overview}

\begin{figure}
\centering
\includegraphics[width=0.8\textwidth]{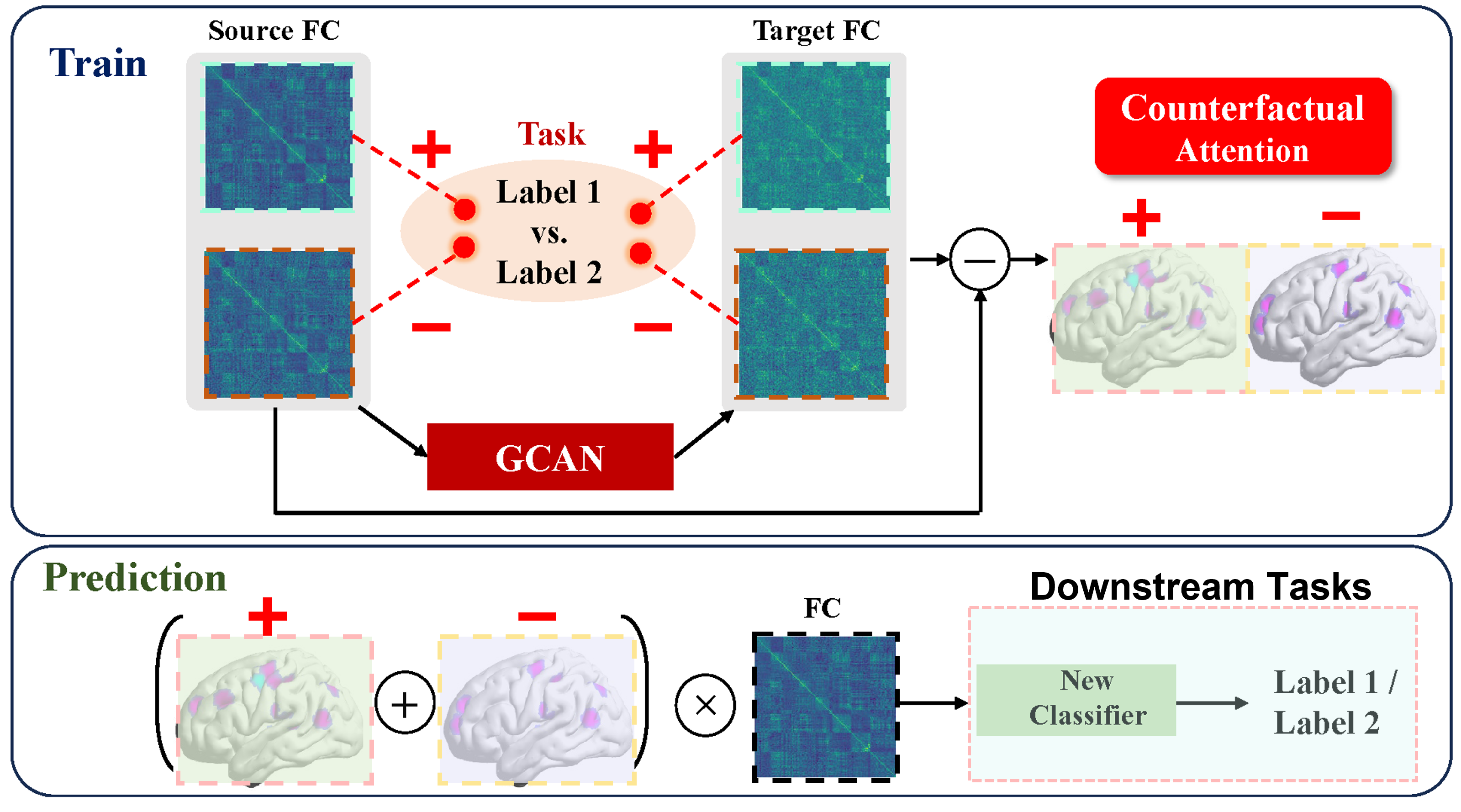}
\caption{Overview of the proposed GCAN framework for counterfactual attention-guided cognitive decline diagnosis. During training, GCAN generates target-label functional connectivity from source-label functional connectivity and derives positive and negative counterfactual attention maps by measuring their differences. During prediction, the learned counterfactual attention is applied to the input functional connectivity to guide the classifier toward disease-related brain connections.}
\label{fig:gcan_overview}
\end{figure}

As shown in Fig.~\ref{fig:gcan_overview}, the proposed framework consists of two stages: counterfactual attention generation and attention-guided diagnosis. In the training stage, GCAN learns to generate a target-label connectome from a source-label connectome. The generated target connectome is compared with the original source connectome to obtain counterfactual attention. In the prediction stage, the aggregated counterfactual attention is applied to the input connectome, so that the classifier focuses on disease-relevant connections rather than the entire connectivity matrix. The original MICCAI version focuses on FC-based counterfactual attention, whereas the journal extension further incorporates SC and constructs a multimodal structure--function counterfactual reasoning framework.

\subsection{Problem formulation}

Let $C \in \mathbb{R}^{N \times N}$ denote a functional connectivity matrix, where $N$ is the number of brain regions defined by an atlas. Given a source-label FC $C_r^s$ with label $y_s$, the goal of GCAN is to generate a target-label FC $C_g^t$ corresponding to label $y_t$:
\begin{equation}
C_g^t = G(C_r^s, y_s, y_t),
\label{eq:uni_target_generation}
\end{equation}
where $G(\cdot)$ denotes the counterfactual generator. The counterfactual attention is defined as the connection-level difference between the generated target FC and the real source FC:
\begin{equation}
A = C_g^t - C_r^s.
\label{eq:uni_attention}
\end{equation}
Positive values in $A$ indicate connections that need to be enhanced when moving from the source state to the target state, while negative values indicate connections that need to be suppressed. The original FC can then be reweighted by the counterfactual attention:
\begin{equation}
\tilde{C}= C \odot A,
\label{eq:uni_masked_fc}
\end{equation}
where $\odot$ denotes element-wise multiplication. In practice, attention maps from different source--target pairs can be aggregated to obtain task-specific positive and negative attention patterns.

For multimodal structure--function modeling, each subject is represented by FC and SC:
\begin{equation}
X^s = [C_{FC}^{s}, C_{SC}^{s}],
\label{eq:multi_input}
\end{equation}
where $C_{FC}^{s}, C_{SC}^{s} \in \mathbb{R}^{N \times N}$. The multimodal generator produces target-label FC and SC:
\begin{equation}
\hat{X}^{t}=G(X^s,y_s,y_t)=[\hat{C}_{FC}^{t},\hat{C}_{SC}^{t}].
\label{eq:multi_target}
\end{equation}
For each modality, counterfactual attention is derived from both forward and reverse cognitive-state transformations. Let $m\in\{FC,SC\}$ denote the modality. Given a source-state connectome $C_m^s$ and a target-state connectome $C_m^t$, the forward counterfactual attention is defined as
\begin{equation}
A_m^{s\rightarrow t}=\hat{C}_m^{t|s}-C_m^s,
\label{eq:forward_attention}
\end{equation}
where $\hat{C}_m^{t|s}$ denotes the generated target-label connectome from the source-label input. Similarly, the reverse counterfactual attention is defined as
\begin{equation}
A_m^{t\rightarrow s}=\hat{C}_m^{s|t}-C_m^t,
\label{eq:reverse_attention}
\end{equation}
where $\hat{C}_m^{s|t}$ denotes the generated source-label connectome from the target-label input.

To avoid cancellation between opposite signed changes, the bidirectional counterfactual attention for each modality is obtained by aggregating the absolute forward and reverse attention maps:
\begin{equation}
A_m = \left|A_m^{s\rightarrow t}\right|+\left|A_m^{t\rightarrow s}\right|,
\quad m\in\{FC,SC\}.
\label{eq:bidirectional_attention}
\end{equation}

The attention-guided FC and SC inputs are then computed separately:
\begin{equation}
\tilde{C}_{FC}=C_{FC}\odot A_{FC}, \qquad
\tilde{C}_{SC}=C_{SC}\odot A_{SC},
\label{eq:modality_specific_masking}
\end{equation}
where $\odot$ denotes element-wise multiplication. The final multimodal input is formed as
\begin{equation}
\tilde{X}=[\tilde{C}_{FC},\tilde{C}_{SC}],
\label{eq:multimodal_input}
\end{equation}
which preserves modality-specific counterfactual attention while allowing the downstream classifier to jointly learn functional and structural discriminative representations.

\subsection{Generative counterfactual attention-guided network}

\begin{figure}
\centering
\includegraphics[width=\textwidth]{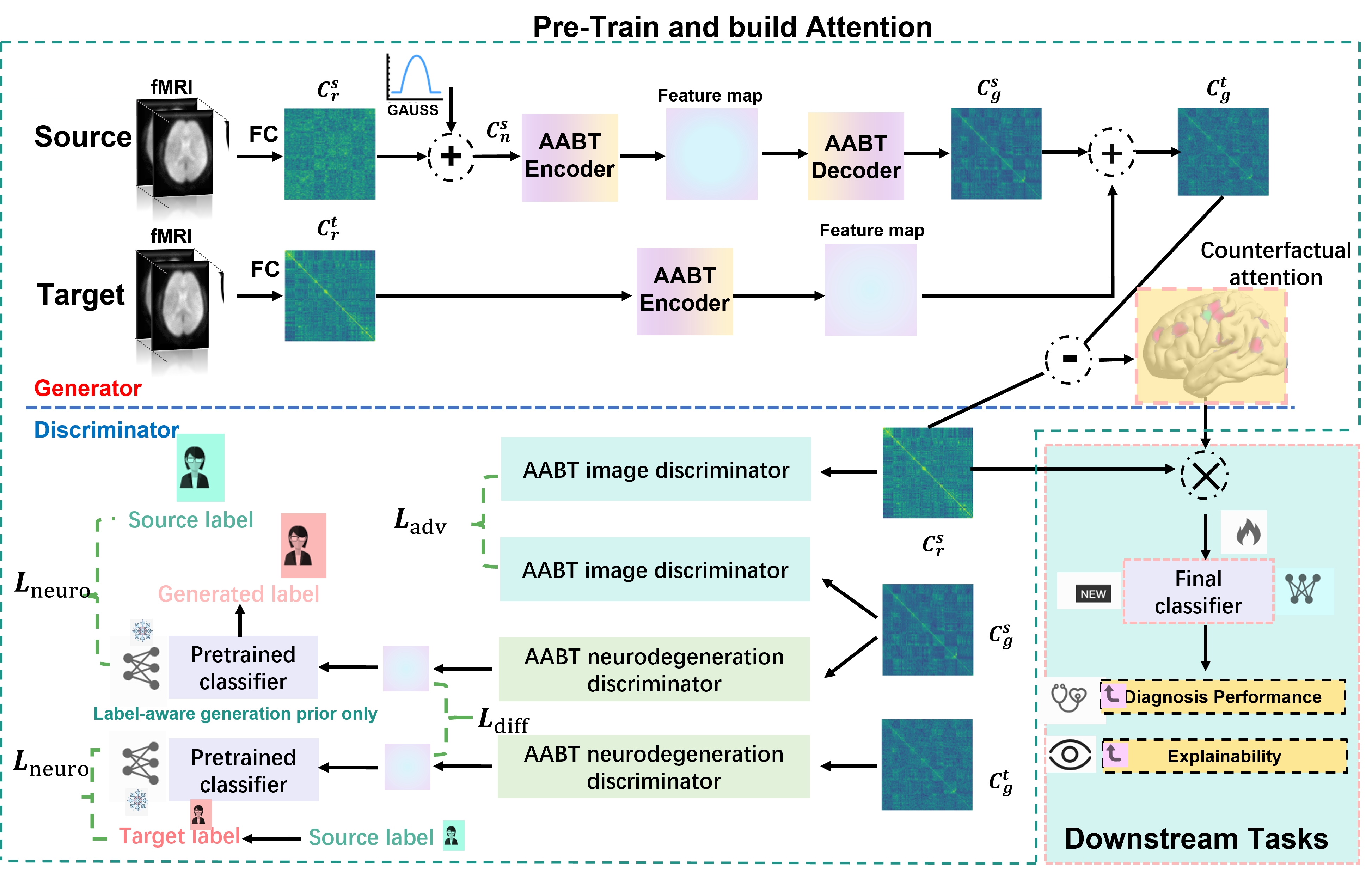}
\caption{Generator--discriminator architecture of the single-modal GCAN. The generator reconstructs the source-label functional connectivity and transforms it into target-label functional connectivity by incorporating target-state information. The discriminator contains image-level and neurodegeneration-level components to constrain the realism and class-discriminative properties of the generated connectomes.}
\label{fig:gcan_gan}
\end{figure}

The detailed generator--discriminator architecture of the single-modal GCAN is illustrated in Fig.~\ref{fig:gcan_gan}.

\subsubsection{Generator.}
The generator contains two steps: source-domain reconstruction and target-domain transformation. First, Gaussian noise is added to the source-label FC to construct a perturbed input:
\begin{equation}
C_n^{s}=C_r^{s}+\epsilon, \qquad \epsilon \sim \mathcal{N}(0,\sigma^2).
\label{eq:noise_fc}
\end{equation}
The reconstruction module maps the noisy source FC back to a source-like generated FC:
\begin{equation}
C_g^{s}=G_{rec}(C_n^{s}).
\label{eq:source_reconstruction}
\end{equation}
Then, the target-label statistical representation is introduced:
\begin{equation}
C_r^{t}=\mathbb{E}[C\mid y=y_t],
\label{eq:target_mean}
\end{equation}
where $C_r^{t}$ is the mean FC of the target class in the training set. The target transformation module generates the target-label FC:
\begin{equation}
C_g^{t}=G_{trans}(C_g^{s},C_r^{t}).
\label{eq:target_transformation}
\end{equation}
This process can be interpreted as learning a source-to-target cognitive shift in the connectivity space, where the difference between $C_g^t$ and $C_r^s$ forms the counterfactual attention.

The generator is optimized using reconstruction, classification, and perceptual constraints:
\begin{equation}
L_G=\lambda_1 L_{rec}+\lambda_2 L_{cls}+\lambda_3 L_{perc},
\label{eq:generator_loss}
\end{equation}
where
\begin{equation}
L_{rec}=\|C_g^{s}-C_r^{s}\|_2^2,
\label{eq:rec_loss}
\end{equation}
\begin{equation}
L_{cls}=CE(f(C_g^{t}),y_t),
\label{eq:cls_loss}
\end{equation}
and
\begin{equation}
L_{perc}=\|\phi(C_g^{s})-\phi(C_r^{s})\|_2^2.
\label{eq:perc_loss}
\end{equation}
Here, $f(\cdot)$ is a pretrained classifier, $\phi(\cdot)$ denotes a high-level feature extractor, $CE(\cdot)$ is the cross-entropy loss, and $\lambda_1$, $\lambda_2$, and $\lambda_3$ are balancing coefficients.

\subsubsection{Discriminator.}
The discriminator contains an image discriminator and a neurodegeneration discriminator. The image discriminator encourages the generated FC to preserve realistic connectivity structures, while the neurodegeneration discriminator encourages generated source and target FCs to contain label-consistent cognitive information. The adversarial loss is defined as
\begin{equation}
L_{adv}=\mathbb{E}[\log D(C_r^s)]+\mathbb{E}[\log(1-D(C_g^s))].
\label{eq:adv_loss}
\end{equation}
The neurodegeneration classification loss is
\begin{equation}
L_{neuro}=CE(f(D(C_g^s)),y_s)+CE(f(D(C_g^t)),y_t).
\label{eq:neuro_loss}
\end{equation}
To further separate source-like and target-like generated representations, a difference constraint is used:
\begin{equation}
L_{diff}=\|D(C_g^t)-D(C_g^s)\|_2^2.
\label{eq:diff_loss}
\end{equation}
The discriminator objective is
\begin{equation}
L_D=L_{adv}+\lambda_4 L_{neuro}+\lambda_5 L_{diff}.
\label{eq:discriminator_loss}
\end{equation}
The overall single-modal objective is
\begin{equation}
L=L_G+L_D.
\label{eq:single_total_loss}
\end{equation}

\subsection{Atlas-aware Bidirectional Transformer}

\begin{figure}
\centering
\includegraphics[width=0.9\textwidth]{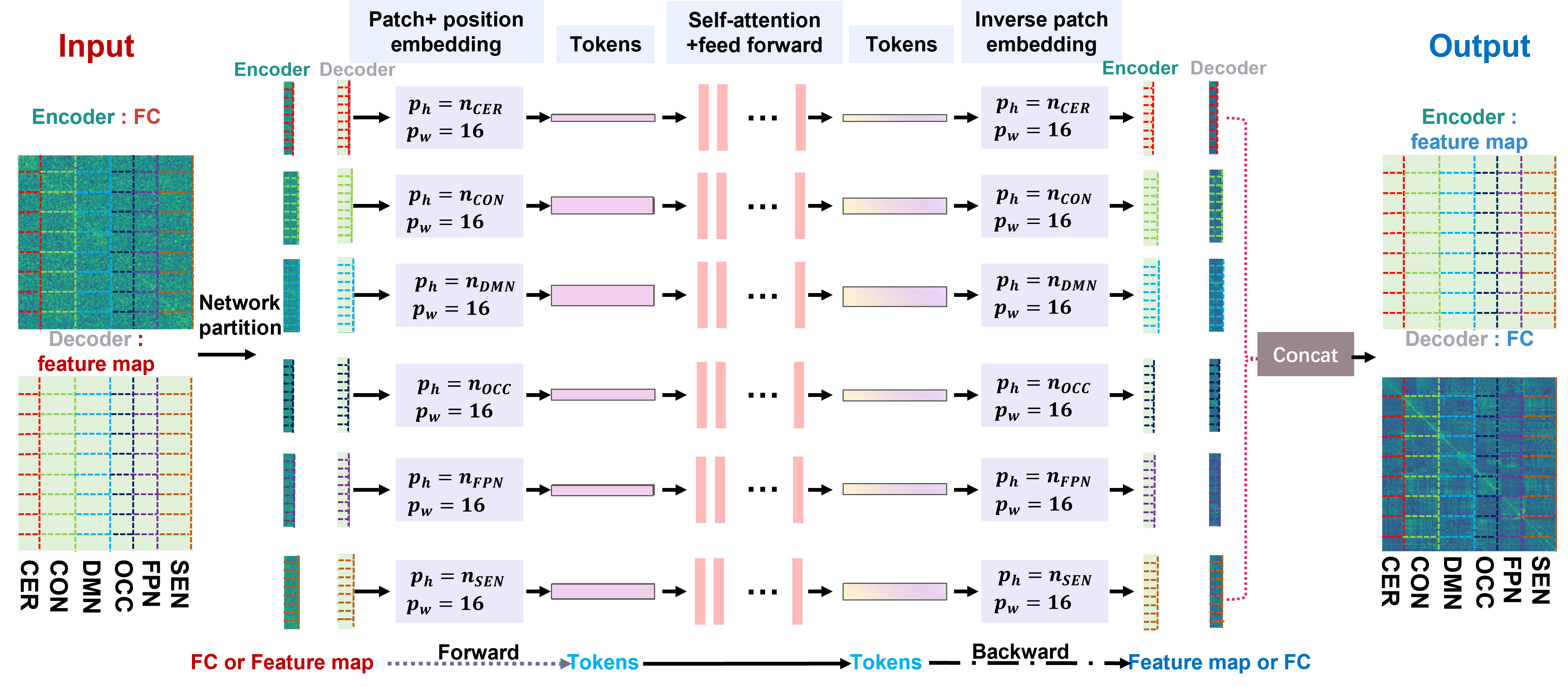}
\caption{Architecture of the AABT for functional connectivity modeling. AABT decomposes the input functional connectivity or feature map into atlas-defined brain networks, performs network-specific patch and position embedding, and then reconstructs the output through inverse patch embedding. This bidirectional encoding--decoding process preserves the structured topology of brain connectomes.}
\label{fig:aabt}
\end{figure}

As shown in Fig.~\ref{fig:aabt}, AABT performs network-level tokenization and reconstruction according to the atlas-defined functional networks. The key difficulty in connectome counterfactual generation lies in preserving the structured properties of FC and SC. Directly feeding the entire connectivity matrix into a Transformer may ignore the atlas-defined organization of brain networks. To address this issue, we design AABT, which divides the connectome into atlas-defined subnetworks and performs network-wise token encoding and inverse token decoding.

Let the connectome be divided into $K$ subnetworks:
\begin{equation}
C=\bigcup_{k=1}^{K} C^{(k)}.
\label{eq:network_partition}
\end{equation}
For the $k$-th subnetwork, a patch embedding operation maps the corresponding connectivity block into tokens:
\begin{equation}
Z^{(k)}=P_k(C^{(k)})+E_k,
\label{eq:patch_embedding}
\end{equation}
where $P_k(\cdot)$ denotes the network-specific patch embedding and $E_k$ denotes the positional embedding. The token representation is then encoded by self-attention and feed-forward layers:
\begin{equation}
\tilde{Z}^{(k)}=\mathrm{FFN}(\mathrm{SA}(Z^{(k)})).
\label{eq:self_attention}
\end{equation}
The inverse patch embedding maps the encoded tokens back to the connectivity space:
\begin{equation}
\hat{C}^{(k)}=P_k^{-1}(\tilde{Z}^{(k)}).
\label{eq:inverse_patch}
\end{equation}
Finally, all reconstructed subnetworks are concatenated:
\begin{equation}
\hat{C}=\mathrm{Concat}(\hat{C}^{(1)},\ldots,\hat{C}^{(K)}).
\label{eq:network_concat}
\end{equation}
In our implementation, the atlas networks include the cerebellum network (CER), cingulo-opercular network (CON), default mode network (DMN), occipital network (OCC), fronto-parietal network (FPN), and sensorimotor network (SEN). The forward patch embedding and backward inverse patch embedding allow AABT to encode and decode connectivity blocks from a global perspective while respecting local network structure.

\subsection{Multimodal structure--function counterfactual reasoning}

\begin{figure}
\centering
\includegraphics[width=0.8\textwidth]{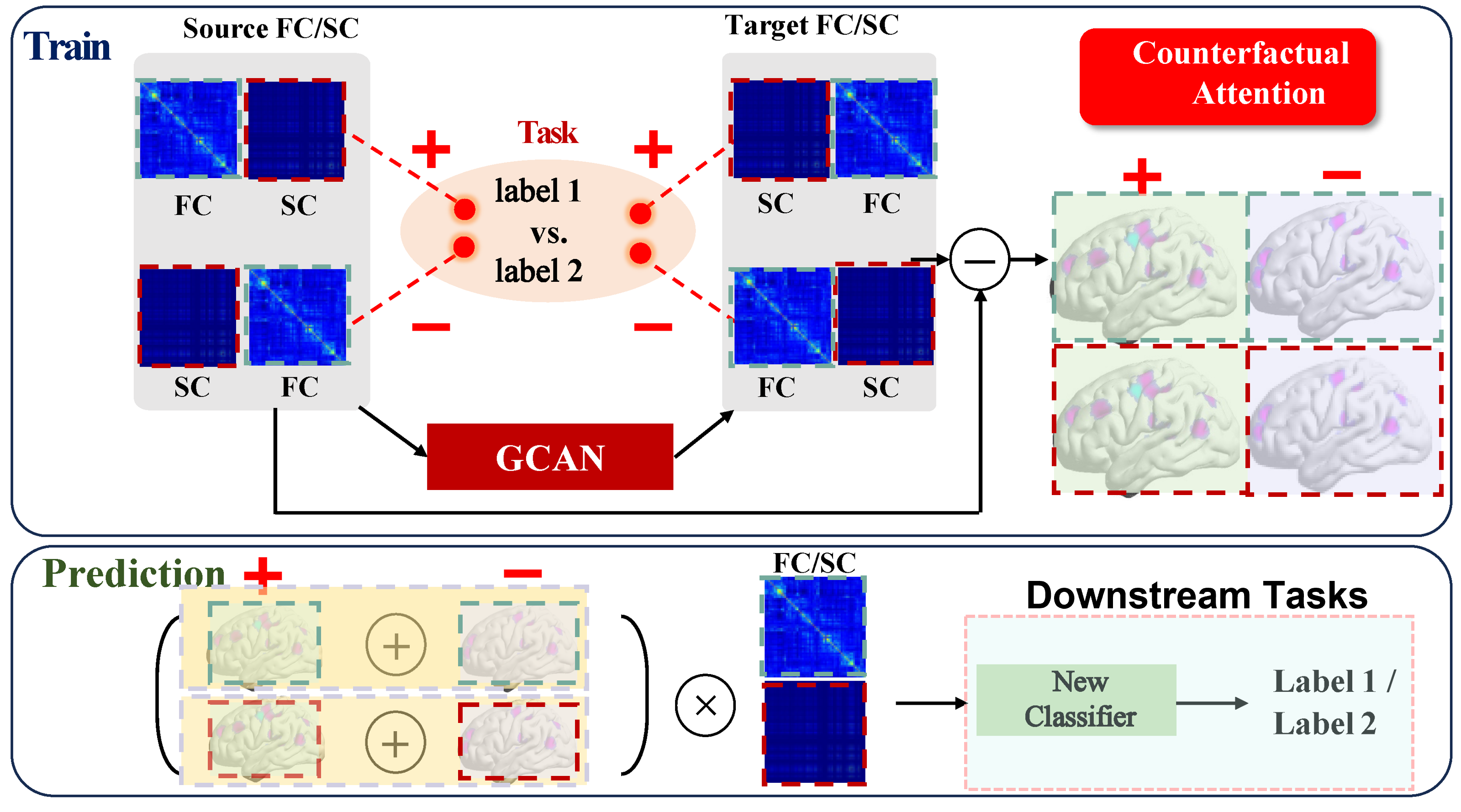}
\caption{Overview of the multimodal GCAN framework for joint functional and structural connectome analysis. In the training stage, source-label FC/SC connectomes are transformed into target-label FC/SC connectomes to construct multimodal counterfactual attention. In the prediction stage, positive and negative attention maps are aggregated and applied to multimodal connectomes to support explainable cognitive decline diagnosis.}
\label{fig:mm_gcan_overview}
\end{figure}

To further model structure--function coupling during cognitive decline, we extend GCAN to a multimodal setting, as shown in Fig.~\ref{fig:mm_gcan_overview}. To extend GCAN from FC-only explanation to structure--function joint explanation, we construct a multimodal generator that takes FC, SC, and their feature maps as inputs. For the $k$-th network, the multimodal input is defined as
\begin{equation}
X_k^{s}=[C_{FC,k}^{s},C_{SC,k}^{s},F_{FC,k}^{s},F_{SC,k}^{s}],
\label{eq:multi_network_input}
\end{equation}
where $F_{FC,k}^{s}$ and $F_{SC,k}^{s}$ denote network-level feature maps corresponding to FC and SC. The cross-modal AABT maps the multimodal input into tokens:
\begin{equation}
Z_k^{s}=P_k(X_k^{s})+E_k.
\label{eq:cross_modal_token}
\end{equation}
The tokens are processed through self-attention and feed-forward modules:
\begin{equation}
\tilde{Z}_k^{s}=\mathrm{FFN}(\mathrm{SA}(Z_k^{s})).
\label{eq:cross_modal_attention}
\end{equation}
The inverse embedding reconstructs multimodal outputs:
\begin{equation}
\hat{X}_k^{s}=P_k^{-1}(\tilde{Z}_k^{s}),
\label{eq:cross_modal_inverse}
\end{equation}
and the complete multimodal representation is obtained by concatenating all network outputs:
\begin{equation}
\hat{X}^{s}=\mathrm{Concat}(\hat{X}_1^{s},\hat{X}_2^{s},\ldots,\hat{X}_K^{s}).
\label{eq:cross_modal_concat}
\end{equation}

\begin{figure}
\includegraphics[width=\textwidth]{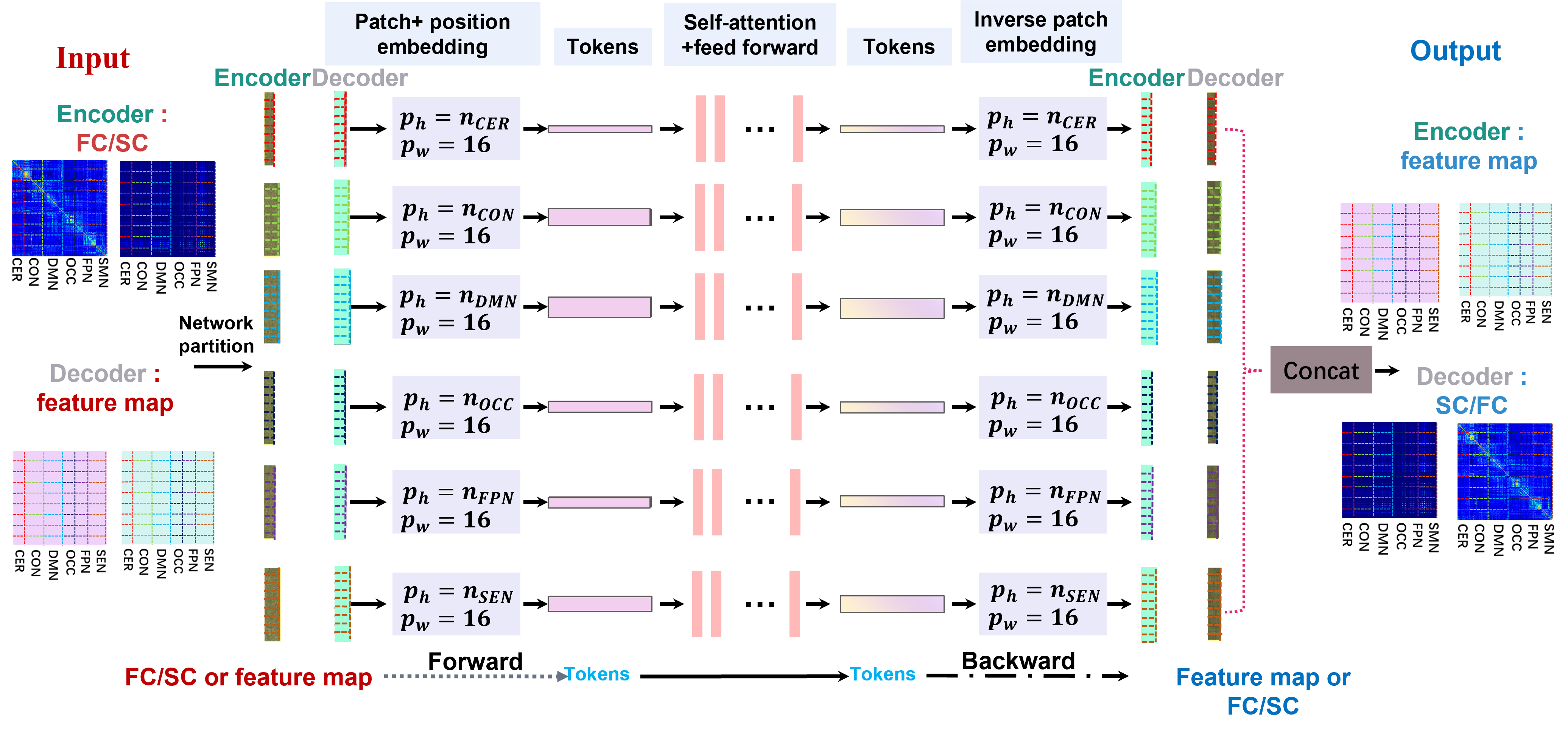}
\caption{Architecture of the cross-modal AABT for joint FC/SC modeling. The input FC, SC, or feature maps are segmented according to atlas-defined brain networks. Network-specific tokens are generated through patch and position embedding, processed by self-attention and feed-forward layers, and then reconstructed through inverse patch embedding to preserve cross-modal connectome topology.}
\label{fig:mm_aabt}
\end{figure}

The cross-modal AABT shown in Fig.~\ref{fig:mm_aabt} enables network-wise encoding and decoding of FC, SC, and their feature maps under atlas constraints.

In the multimodal generation stage, noise is first added to the source FC--SC input:
\begin{equation}
X_n^{s}=X^{s}+\epsilon, \qquad \epsilon\sim \mathcal{N}(0,\sigma^2).
\label{eq:multi_noise}
\end{equation}
The generator reconstructs the source multimodal connectome:
\begin{equation}
\hat{X}^{s}=G_{rec}(X_n^{s},F^{s}),
\label{eq:multi_source_rec}
\end{equation}
and then generates the target-label multimodal connectome using the target-class mean representation:
\begin{equation}
\hat{X}^{t}=G_{trans}(\hat{X}^{s},X_r^{t}), \qquad
X_r^{t}=\mathbb{E}[X\mid y=y_t].
\label{eq:multi_target_gen}
\end{equation}
The FC and SC counterfactual attention maps are then calculated according to Eq.~\eqref{eq:bidirectional_attention}. To visualize region-level changes, the connection-level differences can be mapped to the brain-region space:
\begin{equation}
M=\Psi(\hat{X}^{t})-\Psi(X^{s}),
\label{eq:region_mapping}
\end{equation}
where $\Psi(\cdot)$ denotes the mapping from connectome space to regional activation space. Positive and negative counterfactual maps indicate connection enhancement and suppression during source-to-target transformation.

\begin{figure}
\centering
\includegraphics[width=\textwidth]{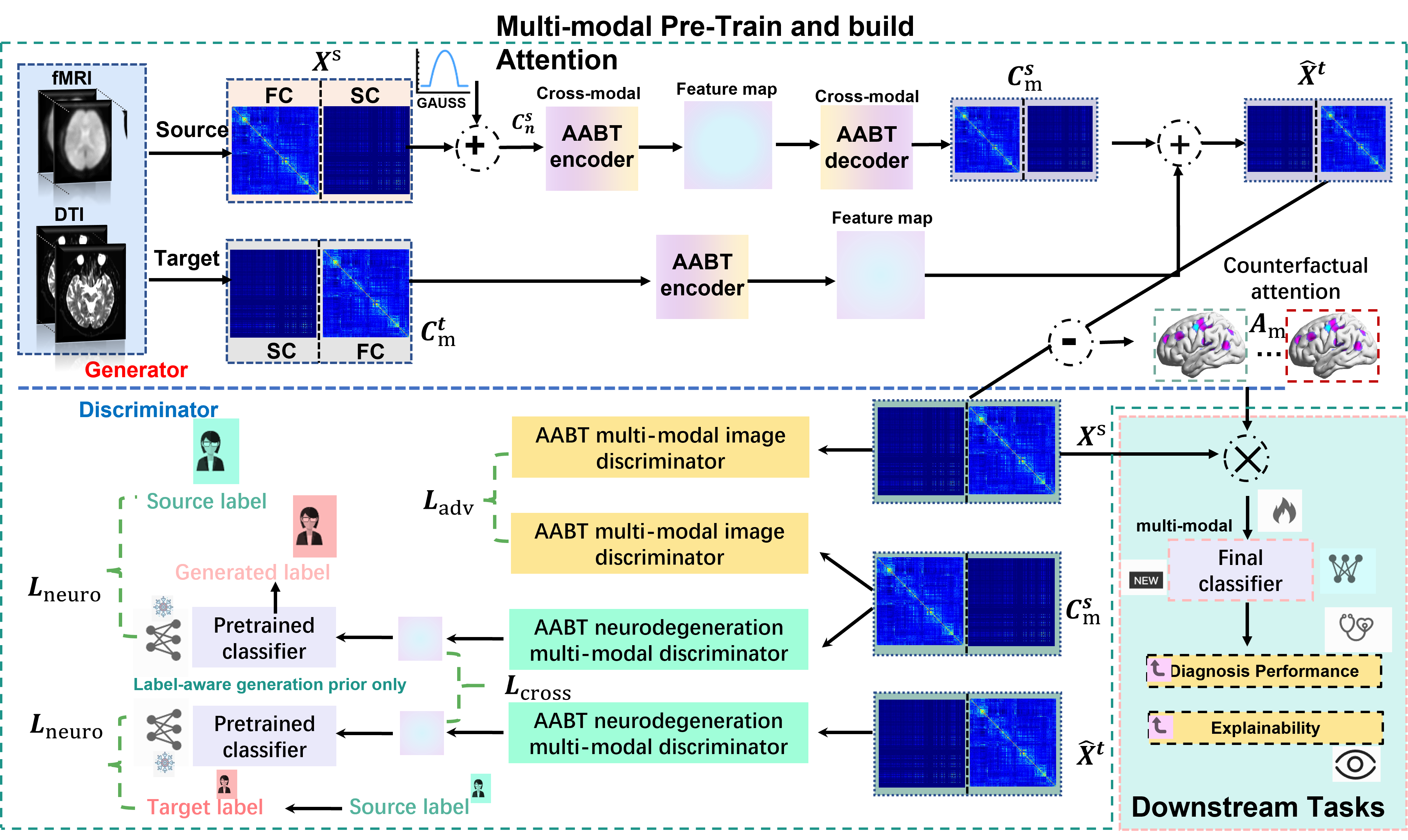}
\caption{Generator--discriminator architecture of the multimodal GCAN. The generator learns source-to-target transformations between multimodal FC/SC connectomes, while the discriminator imposes image-level and neurodegeneration-level constraints on the generated connectomes. Pretrained classifiers provide label-aware supervision for source and target cognitive states.}
\label{fig:mm_gcan_gan}
\end{figure}

The detailed generator--discriminator design of the multimodal GCAN is presented in Fig.~\ref{fig:mm_gcan_gan}.

\subsection{Multimodal optimization}

The multimodal generator is constrained by reconstruction, classification, and perceptual consistency losses:
\begin{equation}
L_{rec}=
\|\hat{C}_{FC}^{s}-C_{FC}^{s}\|_2^2+
\|\hat{C}_{SC}^{s}-C_{SC}^{s}\|_2^2,
\label{eq:multi_rec_loss}
\end{equation}
\begin{equation}
L_{cls}=CE(f(\hat{X}^{t}),y_t),
\label{eq:multi_cls_loss}
\end{equation}
\begin{equation}
L_{perc}=
\|\phi(\hat{X}^{s})-\phi(X^{s})\|_2^2+
\|\phi(\hat{X}^{t})-\phi(X_r^{t})\|_2^2.
\label{eq:multi_perc_loss}
\end{equation}
The multimodal generator loss is
\begin{equation}
L_G=\lambda_1L_{rec}+\lambda_2L_{cls}+\lambda_3L_{perc}.
\label{eq:multi_generator_loss}
\end{equation}
The adversarial loss is
\begin{equation}
L_{adv}=
\mathbb{E}[\log D(X^{s})]+
\mathbb{E}[\log(1-D(\hat{X}^{s}))]+
\mathbb{E}[\log(1-D(\hat{X}^{t}))].
\label{eq:multi_adv_loss}
\end{equation}
The neurodegeneration classification loss is
\begin{equation}
L_{neuro}=CE(f_n(\hat{X}^{s}),y_s)+CE(f_n(\hat{X}^{t}),y_t).
\label{eq:multi_neuro_loss}
\end{equation}
To encourage coordinated but not identical structure--function transformations, we introduce a normalized cross-modal attention consistency loss. Since FC and SC have different value ranges, sparsity levels, and biological meanings, we first normalize the bidirectional counterfactual attention maps:
\begin{equation}
\bar{A}_{m}=
\frac{A_m}{\|A_m\|_F+\epsilon},
\quad m\in\{FC,SC\},
\label{eq:normalized_attention}
\end{equation}
where $A_m$ is the bidirectional counterfactual attention defined in Eq.~\eqref{eq:bidirectional_attention}, $\|\cdot\|_F$ denotes the Frobenius norm, and $\epsilon$ is a small constant for numerical stability. The cross-modal consistency loss is then defined as
\begin{equation}
L_{cross}=\|\bar{A}_{FC}-\bar{A}_{SC}\|_F^2.
\label{eq:cross_loss}
\end{equation}
This term encourages FC and SC to share disease-transition-related attention patterns in the normalized counterfactual space, while still allowing modality-specific functional and structural alterations.

To enhance the distinction between enhanced and suppressed counterfactual patterns, we further define positive and negative regional maps as
\begin{equation}
M^{+}=\mathrm{ReLU}(M), \qquad
M^{-}=\mathrm{ReLU}(-M),
\label{eq:positive_negative_region_maps}
\end{equation}
where $M$ is the regional counterfactual map defined in Eq.~\eqref{eq:region_mapping}. Instead of minimizing the distance between $M^{+}$ and $M^{-}$, we penalize their overlap:
\begin{equation}
L_{att}=
\frac{
\langle M^{+},M^{-}\rangle
}{
\|M^{+}\|_2\|M^{-}\|_2+\epsilon
}.
\label{eq:att_loss}
\end{equation}
Minimizing this term reduces the spatial overlap between positive and negative regional responses, thereby improving the separability of enhanced and suppressed disease-related patterns.

The multimodal discriminator loss is formulated as
\begin{equation}
L_D=L_{adv}+\lambda_4L_{neuro}+\lambda_5L_{cross}+\lambda_6L_{att},
\label{eq:multi_discriminator_loss}
\end{equation}
where $L_{cross}$ promotes normalized FC--SC attention coordination and $L_{att}$ suppresses overlap between positive and negative regional responses.

\section{Experiments}
\label{sec:experiments}


\subsection{Datasets and preprocessing}
\label{sec:datasets_preprocessing}

\begin{table}[htbp]
\caption{Demographic information and rs-fMRI acquisition parameters used in the single-modal FC experiments. Values are presented as mean $\pm$ standard deviation unless otherwise specified.}
\label{tab:single_modal_info}
\centering
\footnotesize
\setlength{\tabcolsep}{4pt}
\renewcommand{\arraystretch}{1.25}

\begin{tabularx}{\textwidth}{
>{\centering\arraybackslash}p{1.35cm}
>{\centering\arraybackslash}p{0.85cm}
>{\centering\arraybackslash}p{0.75cm}
>{\centering\arraybackslash}X
>{\centering\arraybackslash}X
>{\centering\arraybackslash}X
>{\centering\arraybackslash}X
>{\centering\arraybackslash}X}
\toprule
\textbf{Dataset} & \textbf{Group} & \textbf{N} & \textbf{Age (years)} & \textbf{MMSE} & \textbf{MoCA} & \textbf{Education (years)} & \textbf{Weight (kg)} \\
\midrule
GUTCM & HC   & 77 & $64.48 \pm 5.73$ & $29.55 \pm 0.72$ & $26.50 \pm 2.10$ & $12.15 \pm 2.94$ & $85.30 \pm 5.34$ \\
      & SCD  & 75 & $65.24 \pm 5.56$ & $27.00 \pm 0.87$ & $23.20 \pm 2.55$ & $11.04 \pm 2.80$ & $85.60 \pm 5.41$ \\
      & MCI  & 99 & $65.31 \pm 6.70$ & $25.31 \pm 1.04$ & $21.06 \pm 2.75$ & $10.45 \pm 2.94$ & $84.28 \pm 5.02$ \\
\addlinespace[5pt]
ADNI  & HC   & 67 & $76.70 \pm 6.47$ & $29.00 \pm 1.41$ & -- & -- & $79.29 \pm 13.09$ \\
      & SCD  & 22 & $76.21 \pm 5.08$ & -- & -- & -- & $79.89 \pm 15.23$ \\
      & MCI  & 95 & $77.40 \pm 7.07$ & $27.56 \pm 1.89$ & -- & -- & $79.10 \pm 13.99$ \\
\bottomrule
\end{tabularx}

\vspace{5mm}

\begin{tabularx}{\textwidth}{
>{\centering\arraybackslash}p{1.35cm}
>{\centering\arraybackslash}p{1.15cm}
>{\centering\arraybackslash}X
>{\centering\arraybackslash}X
>{\centering\arraybackslash}X
>{\centering\arraybackslash}X
>{\centering\arraybackslash}X
>{\centering\arraybackslash}X}
\toprule
\textbf{Dataset} & \textbf{Modality} & \textbf{TR (ms)} & \textbf{TE (ms)} & \textbf{FOV (mm)} & \textbf{Slice Thickness (mm)} & \textbf{Flip Angle} & \textbf{Slices} \\
\midrule
GUTCM & rs-fMRI & 2000 & 30 & $100 \times 100$ & 5.0 & 90$^\circ$ & 31 \\
ADNI  & rs-fMRI & 2000 & 30  & $93 \times 93$   & 3.3 & 80$^\circ$ & 31/48 \\
\bottomrule
\end{tabularx}

\vspace{1mm}
\begin{flushleft}
\footnotesize
\textit{Note}: ADNI SCD corresponds to the SMC label in the uploaded ADNI file. ``--'' indicates that the corresponding variable was unavailable for the selected subset.
\end{flushleft}
\end{table}

We evaluate the proposed method under both single-modal FC and multimodal FC--SC settings. For the single-modal FC experiments, two datasets are used, including a hospital-collected dataset and the Alzheimer's Disease Neuroimaging Initiative (ADNI) dataset. The hospital-collected dataset, denoted as GUTCM, contains 77 healthy controls (HC), 75 SCD subjects, and 99 MCI subjects. The ADNI dataset contains 67 HC, 22 SCD subjects, and 95 MCI subjects. In ADNI, the SCD group corresponds to subjects labeled as SMC in the original diagnostic records. All rs-fMRI data are preprocessed using SPM12, including slice-timing correction, head-motion estimation and correction, intra-subject registration, and registration to the standard space. The demographic information of the single-modal FC datasets is summarized in Table~\ref{tab:single_modal_info}.

For the multimodal FC--SC experiments, the training process is organized into three stages: FC pre-training, SC pre-training, and paired FC--SC multimodal training. In the FC pre-training stage, GUTCM and ADNI are used together with the SLIM dataset. SLIM contains only HC subjects and is introduced to increase the amount of normative rs-fMRI data for functional-connectivity representation learning. In the SC pre-training stage, ADNI and BJE are used, where BJE also contains only HC subjects and provides additional structural-connectivity information for learning structural priors. Finally, for paired FC--SC multimodal GCAN training, 198 ADNI subjects with both FC and SC data are selected, including 54 HC, 84 SCD, and 60 MCI subjects. This paired subset ensures that functional and structural connectomes are derived from the same subjects during multimodal counterfactual generation. The demographic information of all datasets used in the multimodal training process is summarized in Table~\ref{tab:multimodal_training_info}.


\subsection{MRI acquisition parameters}
\label{sec:mri_parameters}

 Demographic variables include sample size, age, MMSE, MoCA, education years, and body weight when available. Since different datasets provide different metadata fields, unavailable variables are marked as ``--'' in the tables. This design avoids introducing estimated or unavailable demographic values while still presenting the available cohort information.

Table~\ref{tab:single_modal_info} reports the demographic information used in the single-modal FC experiments. Table~\ref{tab:multimodal_training_info} further summarizes the demographic information for the three multimodal training stages, including FC pre-training, SC pre-training, and paired FC--SC multimodal training. This separation is necessary because the pre-training stages use additional auxiliary datasets, whereas the final multimodal GCAN requires paired FC and SC data from the same subjects.

MRI image parameters are summarized in Table~\ref{tab:mri_parameters_all}. The reported parameters include repetition time (TR), echo time (TE), field of view (FOV), slice thickness, flip angle, and number of slices for rs-fMRI and DTI acquisitions. For GUTCM and ADNI, acquisition parameters are reported according to the available protocol information. For BJE and SLIM, FOV, slice thickness, and slice number are derived from available NIfTI header information. Temporal acquisition parameters such as TR, TE, and flip angle are not available for some image files and are therefore marked as ``--''. These heterogeneous imaging protocols provide a realistic setting for evaluating the robustness of the proposed counterfactual connectome modeling framework.

\begin{table}[htbp]
\caption{Demographic information of the datasets used for multimodal training. Values are presented as mean $\pm$ standard deviation unless otherwise specified.}
\label{tab:multimodal_training_info}
\centering
\scriptsize
\setlength{\tabcolsep}{3pt}
\renewcommand{\arraystretch}{1.18}

\begin{tabularx}{\textwidth}{
@{}
>{\centering\arraybackslash}p{1.60cm}
>{\centering\arraybackslash}p{1.05cm}
>{\centering\arraybackslash}p{0.65cm}
>{\centering\arraybackslash}p{0.55cm}
>{\centering\arraybackslash}X
>{\centering\arraybackslash}X
>{\centering\arraybackslash}X
>{\centering\arraybackslash}X
>{\centering\arraybackslash}X
@{}}
\toprule
\textbf{\makecell{Training\\stage}} 
& \textbf{Dataset} 
& \textbf{Group} 
& \textbf{N}
& \textbf{\makecell{Age\\(years)}} 
& \textbf{MMSE} 
& \textbf{MoCA} 
& \textbf{\makecell{Edu.\\(years)}} 
& \textbf{\makecell{Weight\\(kg)}} \\
\midrule

\multirow{7}{*}{\parbox{1.60cm}{\centering FC\\pre-training}}
& GUTCM & HC  & 77  & $64.48 \pm 5.73$ & $29.55 \pm 0.72$ & $26.50 \pm 2.10$ & $12.15 \pm 2.94$ & $85.30 \pm 5.34$ \\
&       & SCD & 75  & $65.24 \pm 5.56$ & $27.00 \pm 0.87$ & $23.20 \pm 2.55$ & $11.04 \pm 2.80$ & $85.60 \pm 5.41$ \\
&       & MCI & 99  & $65.31 \pm 6.70$ & $25.31 \pm 1.04$ & $21.06 \pm 2.75$ & $10.45 \pm 2.94$ & $84.28 \pm 5.02$ \\
\cmidrule(lr){2-9}
& ADNI  & HC  & 67  & $76.70 \pm 6.47$ & $29.00 \pm 1.41$ & -- & -- & $79.29 \pm 13.09$ \\
&       & SCD & 22  & $76.21 \pm 5.08$ & -- & -- & -- & $79.89 \pm 15.23$ \\
&       & MCI & 95  & $77.40 \pm 7.07$ & $27.56 \pm 1.89$ & -- & -- & $79.10 \pm 13.99$ \\
\cmidrule(lr){2-9}
& SLIM  & HC  & 915 & $20.08 \pm 1.28$ & -- & -- & -- & -- \\

\midrule

\multirow{4}{*}{\parbox{1.60cm}{\centering SC\\pre-training}}
& ADNI  & HC  & 94  & $76.58 \pm 6.69$ & $28.86 \pm 1.55$ & -- & -- & $74.29 \pm 18.25$ \\
&       & SCD & 115 & $74.92 \pm 7.70$ & -- & -- & -- & $78.50 \pm 13.09$ \\
&       & MCI & 31  & $76.99 \pm 5.32$ & $27.57 \pm 1.89$ & -- & -- & $79.89 \pm 15.23$ \\
\cmidrule(lr){2-9}
& BJE   & HC  & 180 & $21.22 \pm 1.94$ & -- & -- & -- & -- \\

\midrule

\multirow{3}{*}{\parbox{1.60cm}{\centering FC--SC\\multimodal}}
& ADNI  & HC  & 54 & $78.44 \pm 5.76$ & $29.00 \pm 0.00$ & -- & -- & $65.73 \pm 17.96$ \\
&       & SCD & 84 & $77.07 \pm 5.78$ & -- & -- & -- & $83.06 \pm 17.36$ \\
&       & MCI & 60 & $74.70 \pm 5.56$ & -- & -- & -- & $79.74 \pm 9.81$ \\

\bottomrule
\end{tabularx}

\vspace{1mm}
\begin{flushleft}
\scriptsize
\textit{Note}: HC, healthy control; SCD, subjective cognitive decline; MCI, mild cognitive impairment. ``--'' indicates that the corresponding demographic variable was unavailable.
\end{flushleft}
\end{table}

\begin{table}[htbp]
\centering
\footnotesize
\setlength{\tabcolsep}{4pt}
\renewcommand{\arraystretch}{1.20}

\caption{MRI image parameters of the four datasets used in this study.}
\label{tab:mri_parameters_all}

\begin{tabularx}{\textwidth}{
>{\centering\arraybackslash}p{1.25cm}
>{\centering\arraybackslash}p{1.15cm}
>{\centering\arraybackslash}X
>{\centering\arraybackslash}X
>{\centering\arraybackslash}X
>{\centering\arraybackslash}X
>{\centering\arraybackslash}X
>{\centering\arraybackslash}X}
\toprule
\textbf{Dataset} 
& \textbf{Modality} 
& \textbf{TR (ms)} 
& \textbf{TE (ms)} 
& \textbf{FOV (mm)} 
& \textbf{Slice Thickness (mm)} 
& \textbf{Flip Angle} 
& \textbf{Slices} \\
\midrule

GUTCM & rs-fMRI & 2000 & 30 & $100 \times 100$ & 5.0 & $90^\circ$ & 31 \\
      & DTI     & --   & --  & $90 \times 90$   & 3.0 & -- & 60 \\

\addlinespace[3pt]

ADNI  & rs-fMRI & 2000 & 30  & $93 \times 93$   & 3.3 & $80^\circ$ & 31/48 \\
      & DTI     & --   & --  & $192 \times 192$ & 2.0 & -- & 60 \\

\addlinespace[3pt]

BJE   & rs-fMRI & -- & -- & $200 \times 200$     & 3.6 & -- & 36 \\
      & DTI     & -- & -- & $172.5 \times 172.5$ & 2.5 & -- & 60 \\

\addlinespace[3pt]

SLIM  & rs-fMRI & -- & -- & $220 \times 220$   & 4.0 & -- & 36 \\
      & DTI     & -- & -- & $192 \times 192$   & 2.0 & -- & 60 \\

\bottomrule
\end{tabularx}
\end{table}

\subsection{Implementation details}
\label{sec:implementation_details}

In the generator, the Transformer depth of the encoder and decoder is set to 3. In the image discriminator and neurodegeneration discriminator, the Transformer depth is set to 8. For all Transformer-based modules, the hidden embedding dimension is set to 256, and the number of attention heads is set to 8. The diagnostic classifier is implemented with a ResNet backbone followed by Transformer-based feature modeling. For the main single-modal comparison, ResNet10 with a 16-head Transformer is used as the final diagnostic model. The Adam optimizer is used for model optimization, with an initial learning rate of $1\times10^{-4}$ and a batch size of 16.

All models are implemented in PyTorch and trained on a workstation equipped with two NVIDIA GeForce RTX 4090 GPUs. During training, the generator and discriminator are optimized alternately. The evaluation metrics include accuracy (ACC), recall, precision, and F1-score. For multimodal experiments, we report five-fold cross-validation results using the same metrics.

\subsection{Baseline implementation and fairness}
\label{sec:baseline_fairness}

To ensure a fair comparison, all baseline models were reimplemented under the same preprocessing pipeline, input representation, train/test split, and evaluation metrics whenever possible. Specifically, each baseline was reproduced by following the core methodological novelty described in the corresponding original paper, such as convolutional feature extraction, residual learning, graph-based representation learning, Transformer-based dependency modeling, or hybrid CNN--Transformer design. For methods originally designed for different imaging modalities, feature formats, or datasets, we only adapted the input and output layers to accept the same FC or FC--SC connectome representations used in this study, while preserving their main network architecture and key computational modules.

All reimplemented baselines were trained and evaluated using the same cross-validation protocol as the proposed method. No additional data, labels, or task-specific prior information were introduced for any baseline model. This setting allows the comparison to focus on the effect of the proposed generative counterfactual attention mechanism rather than differences in preprocessing, data partitioning, or evaluation strategy. For methods whose implementation details were not fully specified in the original publications, we followed the reported core design principles and used standard training settings consistent with the rest of the experiments.

\subsection{Data leakage prevention}
\label{sec:data_leakage_prevention}

 Since pre-trained FC and SC classifiers are introduced in the multimodal counterfactual generation stage, we explicitly clarify the strategy used to prevent data leakage. The pre-trained classifiers are used only to provide modality-specific diagnostic priors for target-label connectome generation. They guide the generator to synthesize target-label FC/SC matrices and to derive counterfactual attention maps, but they are not used as the downstream diagnostic models for final performance evaluation.

The downstream diagnostic classifier is trained and evaluated independently under the predefined five-fold cross-validation protocol. Specifically, the prediction scores or classification outputs of the pre-trained FC/SC classifiers are not used as input features for the final classifier. The generated counterfactual attention maps are used to reweight connectome representations, whereas the final diagnostic performance is obtained from a separately trained classifier within each evaluation setting. Therefore, the pre-training stage and the downstream diagnostic evaluation are functionally separated.

This design avoids information leakage from the pre-trained classifiers to the final diagnostic model. In particular, no test-fold labels, test-fold predictions, or test-fold performance information are introduced into the downstream classifier through the pre-trained FC/SC classifiers. As a result, the reported diagnostic performance reflects the effectiveness of the proposed counterfactual attention mechanism rather than any unintended reuse of test-set information.

For each cross-validation split, target-class mean connectomes, counterfactual attention maps, and downstream diagnostic classifiers are derived only from the training folds. The test fold is used only for final evaluation and is never used to estimate target-state priors or attention maps.

\section{Results}
\label{sec:results}

This section evaluates the proposed GCAN framework from both diagnostic performance and interpretability perspectives. Section~\ref{sec:single_modal_results} presents the FC-based single-modal results, including diagnostic performance comparison in Section~\ref{sec:single_modal_diagnosis}, counterfactual attention visualization in Section~\ref{sec:single_modal_attention}, circular connectome analysis in Section~\ref{sec:single_modal_circular}, connectivity matrix similarity analysis in Section~\ref{sec:single_modal_similarity}, and ablation study in Section~\ref{sec:single_modal_ablation}. These analyses verify whether counterfactual attention can improve FC-based cognitive decline diagnosis and identify biologically meaningful functional connectivity alterations.

Section~\ref{sec:multimodal_results} further evaluates the extended multimodal GCAN under the joint FC--SC setting. Specifically, multimodal diagnostic performance is reported in Section~\ref{sec:multimodal_diagnosis}, counterfactual attention maps are visualized in Section~\ref{sec:multimodal_attention}, and multimodal circular connectome patterns are examined in Section~\ref{sec:multimodal_circular}. To further assess interpretability, CAM-based comparison with Grad-CAM and Score-CAM is provided in Section~\ref{sec:cam_comparison}. Statistical performance distributions are analyzed in Section~\ref{sec:multimodal_statistics}. We also quantify structure--function counterfactual overlap in Section~\ref{sec:sf_overlap}, evaluate FC/SC synthesis quality in Section~\ref{sec:multimodal_similarity}, and conduct multimodal ablation studies in Section~\ref{sec:multimodal_ablation}, including counterfactual-attention ablation and AABT topology-preservation analysis. Together, these experiments assess the diagnostic effectiveness, generation reliability, statistical stability, and interpretability of GCAN across different cognitive decline tasks.

\subsection{Single-modal Results}
\label{sec:single_modal_results}
\subsubsection{Single-modal diagnostic performance}
\label{sec:single_modal_diagnosis}

\begin{table}[htbp]
\centering
\scriptsize
\setlength{\tabcolsep}{1.5pt}
\renewcommand{\arraystretch}{1.05}

\caption{Single-modal diagnostic performance on the hospital-collected and ADNI datasets across three classification tasks. $^{*}$ denotes that counterfactual attention is used.}
\label{tab:single_modal_diagnosis}

\begin{tabularx}{\textwidth}{
@{}
>{\centering\arraybackslash}p{1.15cm}
>{\raggedright\arraybackslash}X
*{8}{>{\centering\arraybackslash}p{0.80cm}}
@{}}
\toprule
\multirow{2}{*}{\textbf{Task}} 
& \multirow{2}{*}{\textbf{Method}} 
& \multicolumn{4}{c}{\textbf{Hospital}} 
& \multicolumn{4}{c}{\textbf{ADNI}} \\
\cmidrule(lr){3-6} \cmidrule(lr){7-10}
& & \textbf{ACC} & \textbf{Rec} & \textbf{Pre} & \textbf{F1}
  & \textbf{ACC} & \textbf{Rec} & \textbf{Pre} & \textbf{F1} \\
\midrule

\multirow{11}{*}{\parbox{1.25cm}{\centering HC vs. SCD}}
& Ramzan et al.~\cite{ramzan2020deep} & 0.800 & 0.400 & 0.800 & 0.533 & 0.667 & 0.605 & 0.663 & 0.633 \\
& Wen et al.~\cite{wen2020transfer} & 0.867 & 0.667 & 0.800 & 0.727 & 0.691 & 0.841 & 0.603 & 0.703 \\
& Zhou et al.~\cite{zhou2022multi} & 0.800 & 0.400 & 0.800 & 0.533 & 0.691 & 0.551 & 0.665 & 0.603 \\
& Gao et al.~\cite{gao2023hybrid} & 0.800 & 0.400 & 0.800 & 0.533 & 0.619 & 0.376 & 0.667 & 0.481 \\
& Tian et al.~\cite{tian2023extensible} & 0.800 & 0.400 & 0.800 & 0.533 & 0.605 & 0.708 & 0.580 & 0.638 \\
& He et al.~\cite{he2024spatiotemporal} & 0.800 & 0.400 & 0.800 & 0.533 & 0.631 & 0.322 & 0.714 & 0.444 \\
& Tang et al.~\cite{tang2024multimodal} & 0.833 & 0.467 & 1.000 & 0.636 & 0.679 & \textbf{0.782} & 0.613 & 0.687 \\
& Kang et al.~\cite{kang2025structural} & 0.800 & 0.400 & 0.800 & 0.533 & 0.642 & 0.482 & 0.659 & 0.556 \\
& Li et al.~\cite{li2025multi} & 0.800 & 0.533 & 0.680 & 0.598 & 0.654 & 0.313 & 0.661 & 0.425 \\
& Zhang et al.~\cite{zhang2025classification} & 0.800 & 0.400 & 0.800 & 0.533 & 0.621 & 0.374 & 0.635 & 0.471 \\
& \textbf{Proposed}$^{*}$ & \textbf{0.933} & \textbf{0.867} & \textbf{1.000} & \textbf{0.929} & \textbf{0.728} & 0.667 & \textbf{0.745} & \textbf{0.704} \\

\midrule

\multirow{11}{*}{\parbox{1.25cm}{\centering HC vs. MCI}}
& Ramzan et al.~\cite{ramzan2020deep} & 0.632 & 0.808 & 0.653 & 0.722 & 0.646 & 0.951 & 0.619 & 0.750 \\
& Wen et al.~\cite{wen2020transfer} & 0.646 & 0.917 & 0.633 & 0.749 & 0.621 & \textbf{0.963} & 0.614 & 0.750 \\
& Zhou et al.~\cite{zhou2022multi} & 0.713 & 0.908 & 0.713 & 0.799 & 0.667 & 0.693 & 0.673 & 0.683 \\
& Gao et al.~\cite{gao2023hybrid} & 0.644 & 0.881 & 0.638 & 0.740 & 0.615 & 0.854 & 0.610 & 0.712 \\
& Tian et al.~\cite{tian2023extensible} & 0.667 & 0.920 & 0.647 & 0.759 & 0.647 & 0.836 & 0.651 & 0.732 \\
& He et al.~\cite{he2024spatiotemporal} & 0.701 & 0.774 & \textbf{0.776} & 0.775 & 0.646 & 0.799 & 0.647 & 0.715 \\
& Tang et al.~\cite{tang2024multimodal} & 0.655 & 0.900 & 0.655 & 0.759 & 0.656 & 0.861 & 0.646 & 0.738 \\
& Kang et al.~\cite{kang2025structural} & 0.621 & 0.893 & 0.624 & 0.735 & 0.635 & 0.674 & \textbf{0.704} & 0.688 \\
& Li et al.~\cite{li2025multi} & 0.644 & \textbf{1.000} & 0.625 & 0.769 & 0.667 & 0.785 & 0.672 & 0.724 \\
& Zhang et al.~\cite{zhang2025classification} & 0.678 & 0.889 & 0.663 & 0.759 & 0.646 & 0.819 & 0.661 & 0.732 \\
& \textbf{Proposed}$^{*}$ & \textbf{0.747} & 0.982 & 0.706 & \textbf{0.821} & \textbf{0.697} & 0.865 & 0.671 & \textbf{0.756} \\

\midrule

\multirow{11}{*}{\parbox{1.25cm}{\centering SCD vs. MCI}}
& Ramzan et al.~\cite{ramzan2020deep} & 0.769 & 1.000 & 0.769 & 0.870 & 0.699 & 0.954 & 0.673 & 0.789 \\
& Wen et al.~\cite{wen2020transfer} & 0.850 & \textbf{1.000} & 0.850 & 0.919 & 0.699 & 0.725 & 0.755 & 0.740 \\
& Zhou et al.~\cite{zhou2022multi} & 0.795 & 1.000 & 0.786 & 0.880 & 0.677 & \textbf{0.983} & 0.652 & 0.784 \\
& Gao et al.~\cite{gao2023hybrid} & 0.769 & 1.000 & 0.769 & 0.870 & 0.720 & 0.786 & 0.794 & 0.790 \\
& Tian et al.~\cite{tian2023extensible} & 0.795 & 1.000 & 0.786 & 0.880 & 0.699 & 0.838 & 0.710 & 0.769 \\
& He et al.~\cite{he2024spatiotemporal} & 0.850 & 1.000 & 0.850 & 0.919 & 0.710 & 0.817 & 0.732 & 0.772 \\
& Tang et al.~\cite{tang2024multimodal} & 0.778 & 0.939 & 0.813 & 0.872 & 0.699 & 0.723 & \textbf{0.761} & 0.742 \\
& Kang et al.~\cite{kang2025structural} & 0.769 & 1.000 & 0.769 & 0.870 & 0.688 & 0.914 & 0.678 & 0.778 \\
& Li et al.~\cite{li2025multi} & 0.821 & 1.000 & 0.805 & 0.892 & 0.667 & 0.634 & 0.759 & 0.691 \\
& Zhang et al.~\cite{zhang2025classification} & 0.769 & 1.000 & 0.769 & 0.870 & 0.699 & 0.851 & 0.694 & 0.765 \\
& \textbf{Proposed}$^{*}$ & \textbf{0.949} & 0.971 & \textbf{0.962} & \textbf{0.966} & \textbf{0.731} & 0.866 & \textbf{0.731} & \textbf{0.792} \\

\bottomrule
\end{tabularx}
\end{table}

We first evaluate the FC-only GCAN on three binary diagnostic tasks: HC vs. SCD, HC vs. MCI, and SCD vs. MCI. The proposed model takes counterfactual-attention-weighted FC as input, while baseline models directly use the original FC. Across the hospital-collected and ADNI datasets, GCAN achieves competitive or superior performance in most metrics.

For HC vs. SCD, GCAN obtains the best overall performance. On the hospital-collected dataset, it achieves an ACC of 0.933 and an F1-score of 0.929, substantially outperforming the baselines. On ADNI, it also achieves the highest ACC, precision, and F1-score. This result suggests that counterfactual attention is particularly useful for detecting subtle functional connectivity alterations at the SCD stage.

For HC vs. MCI, GCAN achieves the highest ACC and F1-score on both datasets. On the hospital-collected dataset, the ACC and F1-score reach 0.747 and 0.821, respectively, with a recall of 0.982. On ADNI, GCAN also achieves the best ACC and F1-score. Although some baselines obtain slightly higher recall or precision on individual settings, GCAN provides a more balanced performance across all metrics.

For SCD vs. MCI, GCAN again obtains the highest ACC, precision, and F1-score on the hospital-collected dataset, reaching 0.949, 0.962, and 0.966, respectively. On ADNI, the proposed method achieves the highest ACC and F1-score, indicating that counterfactual attention can capture fine-grained connectivity differences associated with disease progression from SCD to MCI.

\subsubsection{Single-modal counterfactual attention visualization}
\label{sec:single_modal_attention}

\begin{figure}
\centering
\includegraphics[width=\textwidth]{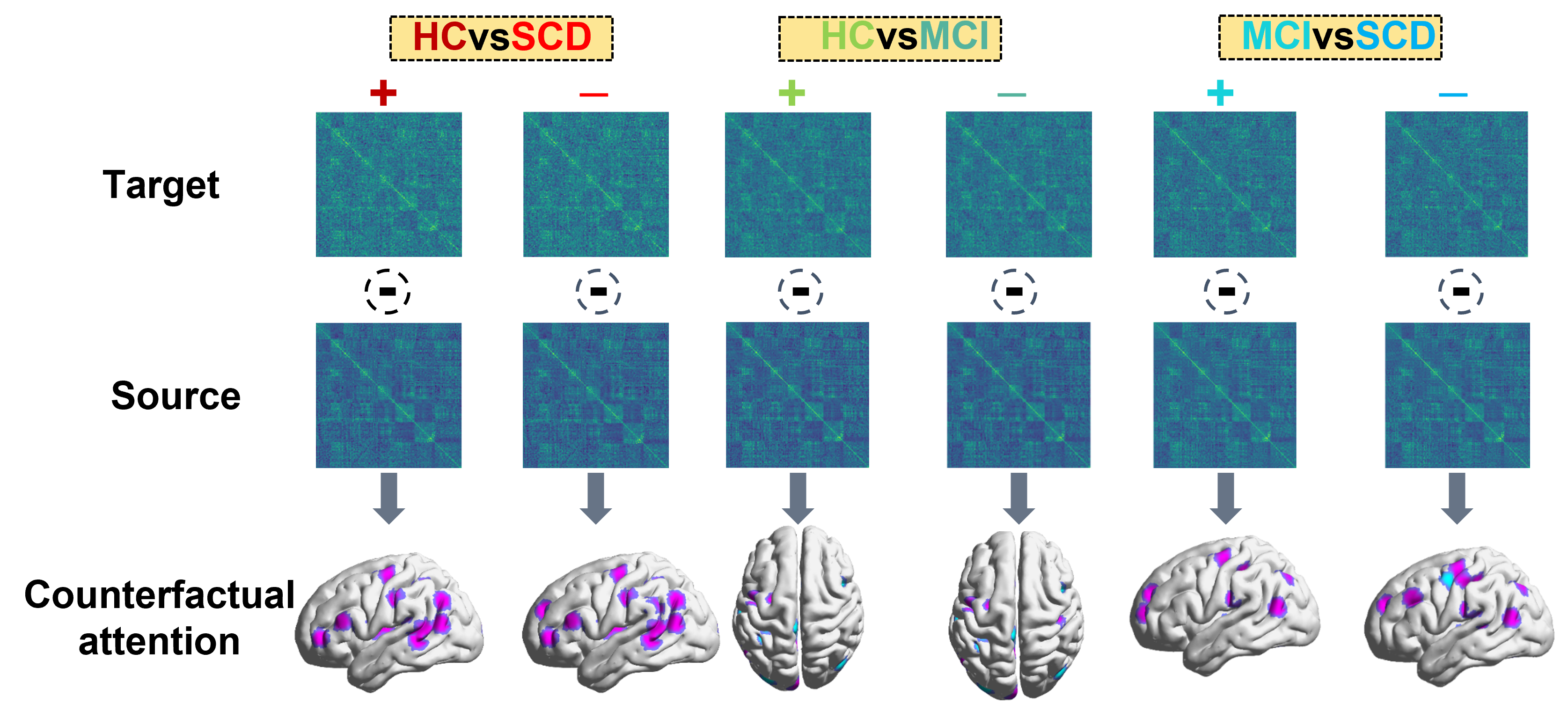}
\caption{Visualization of single-modal counterfactual attention maps across different cognitive-state transition tasks. For each task, the generated target-label FC is compared with the source-label FC, and their difference is projected onto brain regions to obtain positive and negative counterfactual attention maps. The highlighted regions indicate discriminative brain areas associated with functional connectivity changes during cognitive decline.}
\label{fig:single_attention}
\end{figure}

We visualize counterfactual attention maps to examine whether the generated explanations are neurobiologically meaningful. For each diagnostic task, the generated target-label FC and the corresponding counterfactual attention are visualized using BrainNet Viewer, as shown in Fig.~\ref{fig:single_attention}. Across HC vs. SCD, HC vs. MCI, and SCD vs. MCI, the highlighted regions show considerable overlap with networks known to be involved in cognitive decline, including the DMN, FPN, CON, and SEN.

In HC vs. MCI, the attention maps highlight regions such as the prefrontal cortex, cingulate cortex, and hippocampus-related areas, which are consistent with reported FC abnormalities during neurodegeneration. In HC vs. SCD, attention is concentrated on more subtle but discriminative connections, suggesting early functional reorganization before more pronounced impairment. In SCD vs. MCI, the attention maps emphasize differences between high-order cognitive networks, especially DMN--FPN and CON-related connections. These observations indicate that GCAN does not simply enhance the whole FC matrix but identifies sparse and interpretable disease-related connection changes.

The positive and negative attention maps further provide complementary views of the transition between cognitive states. Positive attention reflects connections or regions that become more prominent when the source state is transformed toward the target state, whereas negative attention reflects suppressed or weakened patterns. This bidirectional visualization helps characterize not only where functional abnormalities occur, but also how their direction differs across cognitive-state transitions.

\subsubsection{Single-modal circular connectome analysis}
\label{sec:single_modal_circular}
\begin{figure}
\centering
\includegraphics[width=\textwidth]{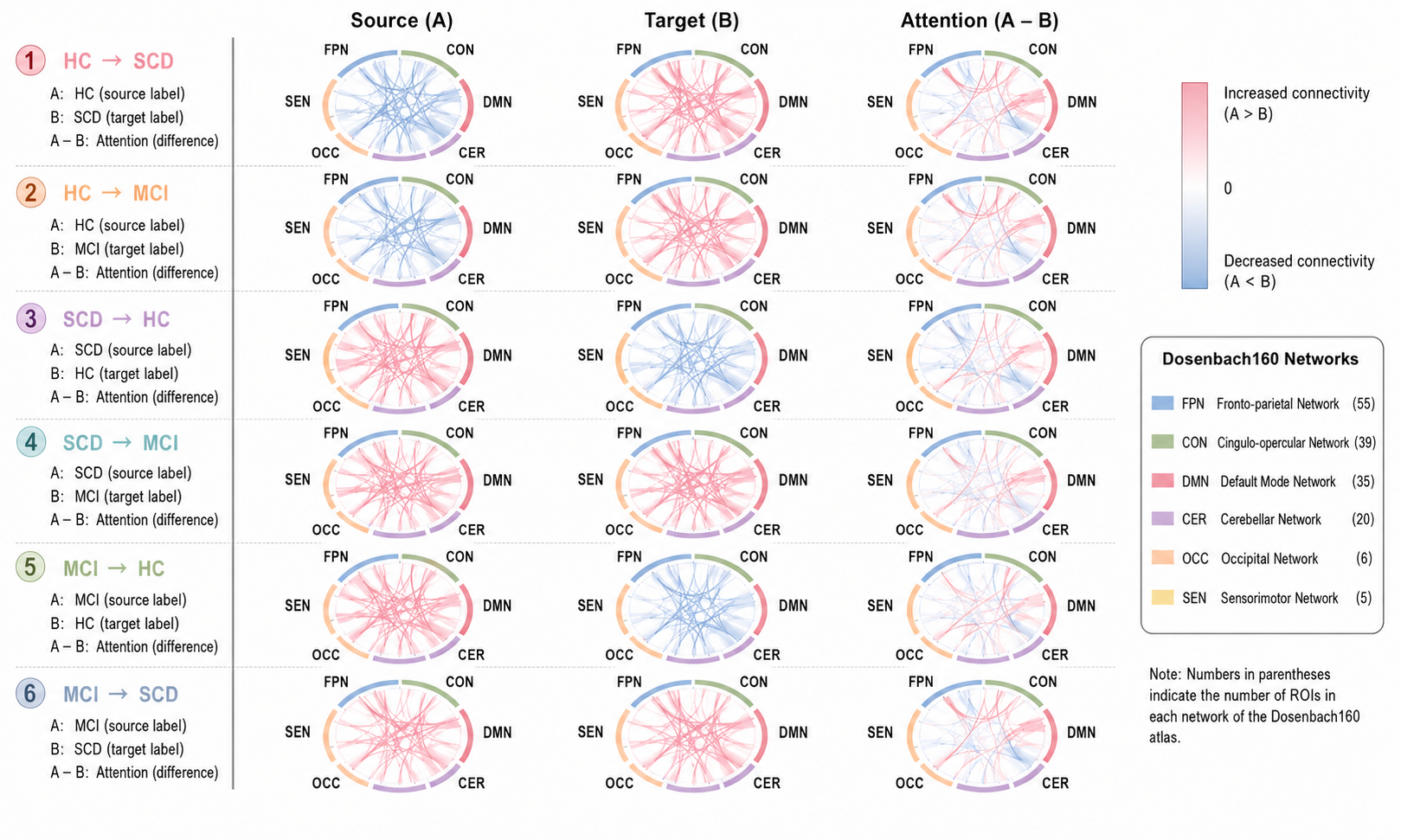}
\caption{Circular connectome visualization of single-modal FC counterfactual analysis across different cognitive-state transition tasks. Source-label FC, generated target-label FC, and counterfactual attention are shown in three columns. Red and blue connections indicate increased and decreased connectivity, respectively.}
\label{fig:connect_uni}
\end{figure}

To further analyze network-level connectivity reorganization, we draw circular connectome diagrams for the source-label FC, generated target-label FC, and counterfactual attention, as shown in Fig.~\ref{fig:connect_uni}. The results show that the differences between source and target states are not uniformly distributed across the brain. Instead, they are concentrated in a limited number of key networks and cross-network connections.

For HC vs. SCD, the counterfactual attention mainly captures early changes involving FPN, DMN, SEN, and CON. For HC vs. MCI, the attention becomes more extensive, indicating that functional abnormalities expand from local network changes to broader cross-network reorganization. For SCD vs. MCI, the most discriminative changes are concentrated in high-order cognitive networks, suggesting that the transition from subjective decline to mild impairment involves subtle but meaningful reconfiguration of functional interactions.

\subsubsection{Single-modal FC synthesis quality analysis}
\label{sec:single_modal_similarity}
\label{sec:single_modal_edgewise_quality}

To evaluate the synthesis quality of single-modal GCAN, we assess the generated target-label FC matrices from both matrix-level and connectome-edge perspectives. Since an FC matrix represents pairwise functional coupling among brain regions, a reliable generated FC should not only resemble the reference matrix at the global structural level, but also preserve relative connectivity patterns and edge-wise connection strengths. Therefore, we use a group of complementary metrics, including PSNR, SSIM, correlation coefficient, MAE, and MSE. Specifically, PSNR measures global intensity-level consistency, SSIM reflects local matrix-structure similarity, and the correlation coefficient evaluates whether the generated FC preserves the relative distribution pattern of functional connections. In addition, MAE and MSE are computed after normalizing all FC matrices to the same value range, providing direct measurements of numerical deviations at the ROI-to-ROI connectivity level. Higher PSNR, SSIM, and correlation coefficients, together with lower MAE and MSE, indicate better preservation of target-state FC patterns. The results are summarized in Table~\ref{tab:single_modal_fc_synthesis_quality}.

\begin{table}[htbp]
\centering
\footnotesize
\renewcommand{\arraystretch}{1.15}
\setlength{\tabcolsep}{5pt}

\caption{Matrix-level similarity, correlation coefficients, and edge-wise reconstruction errors of single-modal FC synthesis under different classification tasks and counterfactual branches.}
\label{tab:single_modal_fc_synthesis_quality}

\begin{tabular*}{\textwidth}{@{\extracolsep{\fill}}llccccc@{}}
\toprule
\multirow{2}{*}{\textbf{Task}} 
& \multirow{2}{*}{\textbf{Branch}} 
& \multicolumn{3}{c}{\textbf{Matrix-level similarity}} 
& \multicolumn{2}{c}{\textbf{Edge-wise error}} \\
\cmidrule(lr){3-5} \cmidrule(lr){6-7}
& & \textbf{PSNR (dB)} & \textbf{SSIM} & \textbf{Corr.}
& \textbf{MAE} & \textbf{MSE} \\
\midrule

\multirow{3}{*}{HC vs. SCD}
& Positive 
& 23.0634 & 0.5529 & 0.4677
& 0.0453 & 0.0049 \\
& Negative 
& 21.9197 & 0.5265 & 0.4077
& 0.0498 & 0.0064 \\
& Average 
& 22.4916 & 0.5397 & 0.4377
& 0.0476 & 0.0057 \\

\midrule

\multirow{3}{*}{HC vs. MCI}
& Positive 
& 18.5253 & 0.5158 & 0.1864
& 0.0576 & 0.0140 \\
& Negative 
& 20.2224 & 0.6580 & 0.3575
& 0.0478 & 0.0095 \\
& Average 
& 19.3739 & 0.5869 & 0.2720
& 0.0527 & 0.0118 \\

\midrule

\multirow{3}{*}{MCI vs. SCD}
& Positive 
& 28.8624 & 0.8026 & 0.8756
& 0.0234 & 0.0013 \\
& Negative 
& 20.1739 & 0.5683 & 0.3236
& 0.0443 & 0.0096 \\
& Average 
& 24.5182 & 0.6855 & 0.5996
& 0.0339 & 0.0055 \\

\midrule

\multirow{3}{*}{Average}
& Positive 
& 23.4837 & 0.6238 & 0.5099
& 0.0421 & 0.0067 \\
& Negative 
& 20.7720 & 0.5843 & 0.3629
& 0.0473 & 0.0085 \\
& All 
& 22.1279 & 0.6040 & 0.4364
& 0.0447 & 0.0076 \\

\bottomrule
\end{tabular*}

\vspace{1mm}
\begin{flushleft}
\footnotesize
\textit{Note}: PSNR, SSIM, and Corr. evaluate matrix-level similarity between paired generated and reference FC matrices. Corr. denotes the correlation coefficient between generated and reference FC matrices. MAE and MSE are computed at the edge level after normalizing all FC matrices to the same value range. Positive and negative branches correspond to the two counterfactual synthesis directions shown in the visualization.
\end{flushleft}
\end{table}

As shown in Table~\ref{tab:single_modal_fc_synthesis_quality}, the generated FC matrices achieve an average PSNR of $22.1279$ dB, SSIM of $0.6040$, correlation coefficient of $0.4364$, normalized MAE of $0.0447$, and normalized MSE of $0.0076$ across all tasks and branches. These results indicate that the generated target-label FC matrices retain a reasonable degree of similarity to the reference FC matrices in terms of global matrix structure, relative connectivity distribution, and edge-wise connection strength. The positive branch generally obtains higher PSNR, SSIM, and correlation values and lower MAE/MSE than the negative branch, suggesting that it produces slightly more consistent FC patterns in the current setting. Among the three tasks, MCI vs. SCD achieves the best overall synthesis quality, with the highest average PSNR, SSIM, and correlation coefficient, as well as the lowest MAE and MSE. This suggests that the generated FC matrices better preserve target-state functional connectivity patterns in this fine-grained transition.

Overall, the combination of these metrics provides a more complete evaluation of FC synthesis quality. PSNR assesses whether the generated matrix preserves the global magnitude distribution, SSIM examines whether local matrix structures are maintained, the correlation coefficient evaluates whether the relative organization of functional connections is consistent with the reference FC, and MAE/MSE quantify the absolute deviations of individual ROI-to-ROI edges after normalization. Therefore, agreement across these complementary indicators supports the reliability of the generated FC matrices for subsequent counterfactual attention computation. Meanwhile, these metrics should be interpreted as quantitative evidence of connectome synthesis fidelity rather than direct proof of biological causality.

\subsubsection{Single-modal ablation study}
\label{sec:single_modal_ablation}

\begin{table}[htbp]
\centering
\footnotesize
\renewcommand{\arraystretch}{1.1}
\caption{Single-modal ablation study on the hospital-collected and ADNI datasets.}
\label{tab:single_ablation}

\begin{tabular*}{\textwidth}{@{\extracolsep{\fill}}llcccccccc@{}}
\toprule
\multirow{2}{*}{\textbf{Task}} 
& \multirow{2}{*}{\textbf{CA}} 
& \multicolumn{4}{c}{\textbf{Hospital-collected dataset}} 
& \multicolumn{4}{c}{\textbf{ADNI dataset}} \\
\cmidrule(lr){3-6} \cmidrule(lr){7-10}
& & \textbf{ACC} & \textbf{Recall} & \textbf{Precision} & \textbf{F1}
  & \textbf{ACC} & \textbf{Recall} & \textbf{Precision} & \textbf{F1} \\
\midrule
HC vs. SCD & \ding{55} & 0.800 & 0.400 & 0.800 & 0.533 & 0.667 & 0.605 & 0.663 & 0.633 \\
HC vs. SCD & \ding{52} & \textbf{0.933} & \textbf{0.867} & \textbf{1.000} & \textbf{0.929} & \textbf{0.728} & \textbf{0.667} & \textbf{0.745} & \textbf{0.704} \\
HC vs. MCI & \ding{55} & 0.632 & 0.808 & 0.653 & 0.722 & 0.646 & \textbf{0.951} & 0.619 & 0.750 \\
HC vs. MCI & \ding{52} & \textbf{0.747} & \textbf{0.982} & \textbf{0.706} & \textbf{0.821} & \textbf{0.697} & 0.865 & \textbf{0.671} & \textbf{0.756} \\
SCD vs. MCI & \ding{55} & 0.769 & \textbf{1.000} & 0.769 & 0.870 & 0.699 & \textbf{0.954} & 0.673 & 0.789 \\
SCD vs. MCI & \ding{52} & \textbf{0.949} & 0.971 & \textbf{0.962} & \textbf{0.966} & \textbf{0.731} & 0.866 & \textbf{0.731} & \textbf{0.792} \\
\bottomrule
\end{tabular*}

\vspace{1mm}
\begin{flushleft}
\footnotesize
\textit{Note}: CA denotes counterfactual attention. \ding{52} indicates that counterfactual attention is used, and \ding{55} indicates that it is not used.
\end{flushleft}
\end{table}

To validate the contribution of counterfactual attention, we compare the same diagnostic model with and without counterfactual attention. The model without counterfactual attention directly uses the original FC, whereas the model with counterfactual attention uses the attention-weighted FC. Across all three tasks and two datasets, introducing counterfactual attention improves most evaluation metrics.

For HC vs. SCD, the improvement is especially clear: on the hospital-collected dataset, ACC increases from 0.800 to 0.933 and F1-score increases from 0.533 to 0.929; on ADNI, ACC and F1-score increase from 0.667 and 0.633 to 0.728 and 0.704. For HC vs. MCI, ACC and F1-score increase from 0.632 and 0.722 to 0.747 and 0.821 on the hospital-collected dataset. For SCD vs. MCI, counterfactual attention improves ACC, precision, and F1-score on both datasets. These results confirm that the proposed attention mechanism enhances discriminative feature learning by focusing on disease-transition-related connections.

\subsection{Multimodal Results}
\label{sec:multimodal_results}

\subsubsection{Multimodal diagnostic performance}
\label{sec:multimodal_diagnosis}

\begin{table}[htbp]
\centering
\scriptsize
\setlength{\tabcolsep}{3pt}
\renewcommand{\arraystretch}{1.08}

\caption{Performance comparison across three classification tasks (\%, five-fold cross-validation). $^{*}$ denotes using counterfactual attention.}
\label{tab:multitask_compare_merge}

\begin{tabular*}{\textwidth}{@{\extracolsep{\fill}}llcccc@{}}
\toprule
\textbf{Task} & \textbf{Method} & \textbf{ACC} & \textbf{Recall} & \textbf{Precision} & \textbf{F1} \\
\midrule

\multirow{12}{*}{HC vs. SCD}
& Ramzan et al.~\cite{ramzan2020deep} & $61.33 \pm 12.93$ & $60.00 \pm 16.16$ & $58.50 \pm 25.77$ & $53.45 \pm 17.83$ \\
& Li et al.~\cite{li2023identification} & $38.67 \pm 2.67$ & $41.67 \pm 10.54$ & $22.00 \pm 2.45$ & $27.86 \pm 1.43$ \\
& Zuo et al.~\cite{zuo2023alzheimer} & $44.00 \pm 29.39$ & $40.00 \pm 25.50$ & $32.00 \pm 29.97$ & $33.57 \pm 25.53$ \\
& Adarsh et al.~\cite{adarsh2024multimodal} & $50.67 \pm 11.62$ & $46.67 \pm 15.46$ & $42.67 \pm 20.37$ & $41.55 \pm 14.30$ \\
& Sibilano et al.~\cite{sibilano2024understanding} & $62.67 \pm 17.18$ & $63.33 \pm 10.00$ & $61.67 \pm 26.01$ & $58.81 \pm 18.70$ \\
& Tang et al.~\cite{tang2024multimodal} & $50.67 \pm 17.18$ & $55.00 \pm 16.33$ & $40.67 \pm 26.00$ & $44.10 \pm 21.63$ \\
& Feng et al.~\cite{feng2025crossmodal} & $58.67 \pm 19.50$ & $60.00 \pm 20.00$ & $47.67 \pm 29.43$ & $50.21 \pm 24.66$ \\
& Huang et al.~\cite{huang2026transformer} & $46.67 \pm 11.16$ & $45.00 \pm 10.00$ & $25.00 \pm 4.47$ & $31.43 \pm 5.13$ \\
\cmidrule(lr){2-6}
& CNN$^{*}$ & $73.33 \pm 8.43$ & $66.67 \pm 13.94$ & $64.33 \pm 26.74$ & $61.98 \pm 19.03$ \\
& Transformer$^{*}$ & $66.67 \pm 18.38$ & $68.33 \pm 9.72$ & $70.83 \pm 27.51$ & $62.38 \pm 19.93$ \\
& CNN+Transformer$^{*}$ & $70.67 \pm 22.55$ & $71.67 \pm 19.44$ & $63.50 \pm 33.52$ & $63.74 \pm 28.01$ \\
& ResNet+Transformer$^{*}$ & $\mathbf{76.00 \pm 23.32}$ & $\mathbf{75.00 \pm 24.72}$ & $\mathbf{77.50 \pm 27.84}$ & $\mathbf{72.62 \pm 27.04}$ \\
\midrule

\multirow{12}{*}{HC vs. MCI}
& Ramzan et al.~\cite{ramzan2020deep} & $50.00 \pm 31.62$ & $50.00 \pm 31.62$ & $40.00 \pm 33.91$ & $43.33 \pm 32.66$ \\
& Li et al.~\cite{li2023identification} & $40.00 \pm 20.00$ & $40.00 \pm 20.00$ & $20.00 \pm 10.00$ & $26.67 \pm 13.33$ \\
& Zuo et al.~\cite{zuo2023alzheimer} & $31.67 \pm 18.56$ & $30.00 \pm 18.71$ & $28.33 \pm 19.44$ & $29.00 \pm 19.08$ \\
& Adarsh et al.~\cite{adarsh2024multimodal} & $46.67 \pm 6.67$ & $45.00 \pm 10.00$ & $35.00 \pm 12.25$ & $38.33 \pm 10.00$ \\
& Sibilano et al.~\cite{sibilano2024understanding} & $40.00 \pm 33.91$ & $40.00 \pm 33.91$ & $31.67 \pm 35.12$ & $34.67 \pm 34.36$ \\
& Tang et al.~\cite{tang2024multimodal} & $41.67 \pm 10.54$ & $45.00 \pm 10.00$ & $26.67 \pm 12.25$ & $32.33 \pm 10.20$ \\
& Feng et al.~\cite{feng2025crossmodal} & $41.67 \pm 10.54$ & $45.00 \pm 10.00$ & $26.67 \pm 12.25$ & $32.33 \pm 10.20$ \\
& Huang et al.~\cite{huang2026transformer} & $40.00 \pm 20.00$ & $40.00 \pm 20.00$ & $20.00 \pm 10.00$ & $26.67 \pm 13.33$ \\
\cmidrule(lr){2-6}
& GCN$^{*}$ & $53.33 \pm 6.67$ & $50.00 \pm 0.00$ & $26.67 \pm 3.33$ & $34.67 \pm 2.67$ \\
& Transformer$^{*}$ & $60.00 \pm 33.91$ & $60.00 \pm 33.91$ & $58.33 \pm 38.73$ & $56.00 \pm 35.18$ \\
& ViT$^{*}$ & $60.00 \pm 33.91$ & $60.00 \pm 33.91$ & $63.33 \pm 35.59$ & $59.33 \pm 33.63$ \\
& ResNet+Transformer$^{*}$ & $\mathbf{66.67 \pm 23.57}$ & $\mathbf{65.00 \pm 22.91}$ & $\mathbf{68.33 \pm 24.49}$ & $\mathbf{65.67 \pm 23.41}$ \\
\midrule

\multirow{12}{*}{SCD vs. MCI}
& Ramzan et al.~\cite{ramzan2020deep} & $54.00 \pm 14.97$ & $55.00 \pm 17.95$ & $50.00 \pm 24.72$ & $48.05 \pm 18.91$ \\
& Li et al.~\cite{li2023identification} & $59.00 \pm 11.14$ & $51.67 \pm 13.33$ & $39.67 \pm 21.92$ & $42.88 \pm 15.61$ \\
& Zuo et al.~\cite{zuo2023alzheimer} & $\mathbf{75.00 \pm 14.83}$ & $\mathbf{70.00 \pm 18.71}$ & $66.17 \pm 30.04$ & $64.90 \pm 24.22$ \\
& Adarsh et al.~\cite{adarsh2024multimodal} & $58.00 \pm 13.27$ & $55.00 \pm 14.53$ & $48.50 \pm 27.09$ & $46.79 \pm 17.87$ \\
& Sibilano et al.~\cite{sibilano2024understanding} & $67.00 \pm 8.72$ & $63.33 \pm 10.00$ & $63.50 \pm 20.71$ & $60.74 \pm 13.78$ \\
& Tang et al.~\cite{tang2024multimodal} & $45.00 \pm 22.36$ & $41.67 \pm 21.73$ & $40.83 \pm 27.94$ & $39.95 \pm 23.33$ \\
& Feng et al.~\cite{feng2025crossmodal} & $45.00 \pm 18.44$ & $40.00 \pm 17.80$ & $35.83 \pm 26.03$ & $36.38 \pm 20.18$ \\
& Huang et al.~\cite{huang2026transformer} & $46.00 \pm 8.00$ & $48.33 \pm 3.33$ & $27.33 \pm 8.07$ & $33.60 \pm 4.62$ \\
\cmidrule(lr){2-6}
& ResNet$^{*}$ & $63.00 \pm 14.00$ & $58.33 \pm 15.81$ & $56.83 \pm 25.99$ & $54.79 \pm 18.98$ \\
& Transformer$^{*}$ & $63.00 \pm 14.00$ & $60.00 \pm 13.33$ & $60.17 \pm 22.54$ & $57.07 \pm 16.16$ \\
& ViT$^{*}$ & $68.00 \pm 24.00$ & $66.67 \pm 25.28$ & $\mathbf{67.50 \pm 28.92}$ & $\mathbf{64.95 \pm 26.56}$ \\
& ResNet+Transformer$^{*}$ & $55.00 \pm 13.42$ & $50.00 \pm 15.81$ & $44.33 \pm 23.11$ & $45.26 \pm 17.76$ \\
\bottomrule
\end{tabular*}
\end{table}

We then evaluate multimodal GCAN using paired FC and SC. Compared with single-modal analysis, multimodal modeling aims to incorporate complementary functional and structural information for cognitive-state discrimination. As shown in Table~\ref{tab:multitask_compare_merge}, the counterfactual-attention-based models generally achieve competitive performance across the three classification tasks. In HC vs. SCD, ResNet+Transformer$^{*}$ obtains the highest average ACC and F1-score among the compared methods, suggesting that multimodal counterfactual attention may help capture subtle early-stage connectivity changes. In HC vs. MCI, ResNet+Transformer$^{*}$ also shows a favorable average performance, although the relatively large standard deviations indicate noticeable fold-wise variability. For SCD vs. MCI, the performance differences among methods are less consistent, reflecting the difficulty of fine-grained discrimination between adjacent cognitive states.

Overall, these results suggest that multimodal counterfactual attention can provide useful complementary information for FC--SC-based diagnosis. However, the relatively large standard deviations, especially in HC vs. MCI and SCD vs. MCI, indicate that the observed improvements should be interpreted cautiously. Further validation with larger paired multimodal cohorts is needed to confirm the robustness of the proposed multimodal extension.

\subsubsection{Multimodal counterfactual attention visualization}
\label{sec:multimodal_attention}
\begin{figure}
\centering
\includegraphics[width=0.9\textwidth]{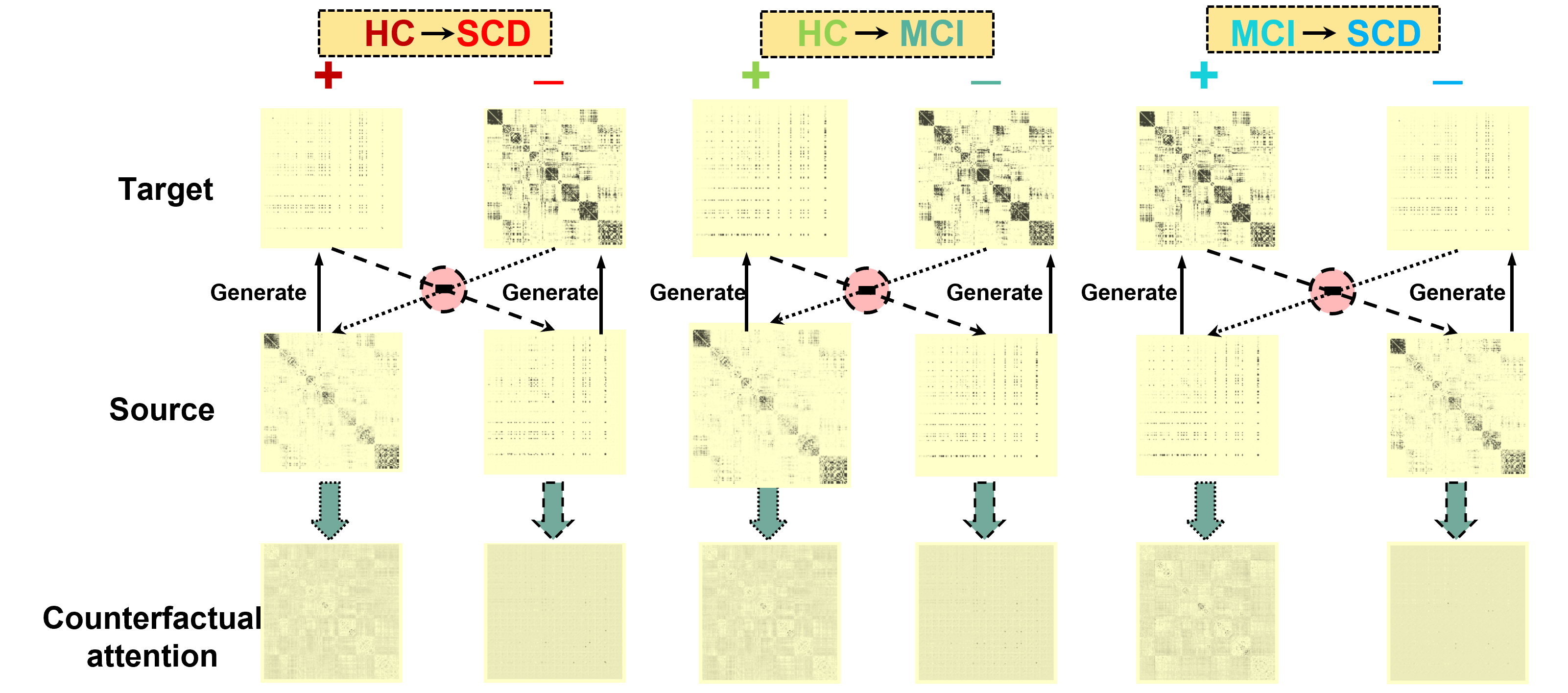}
\vspace{1mm}
\includegraphics[width=0.9\textwidth]{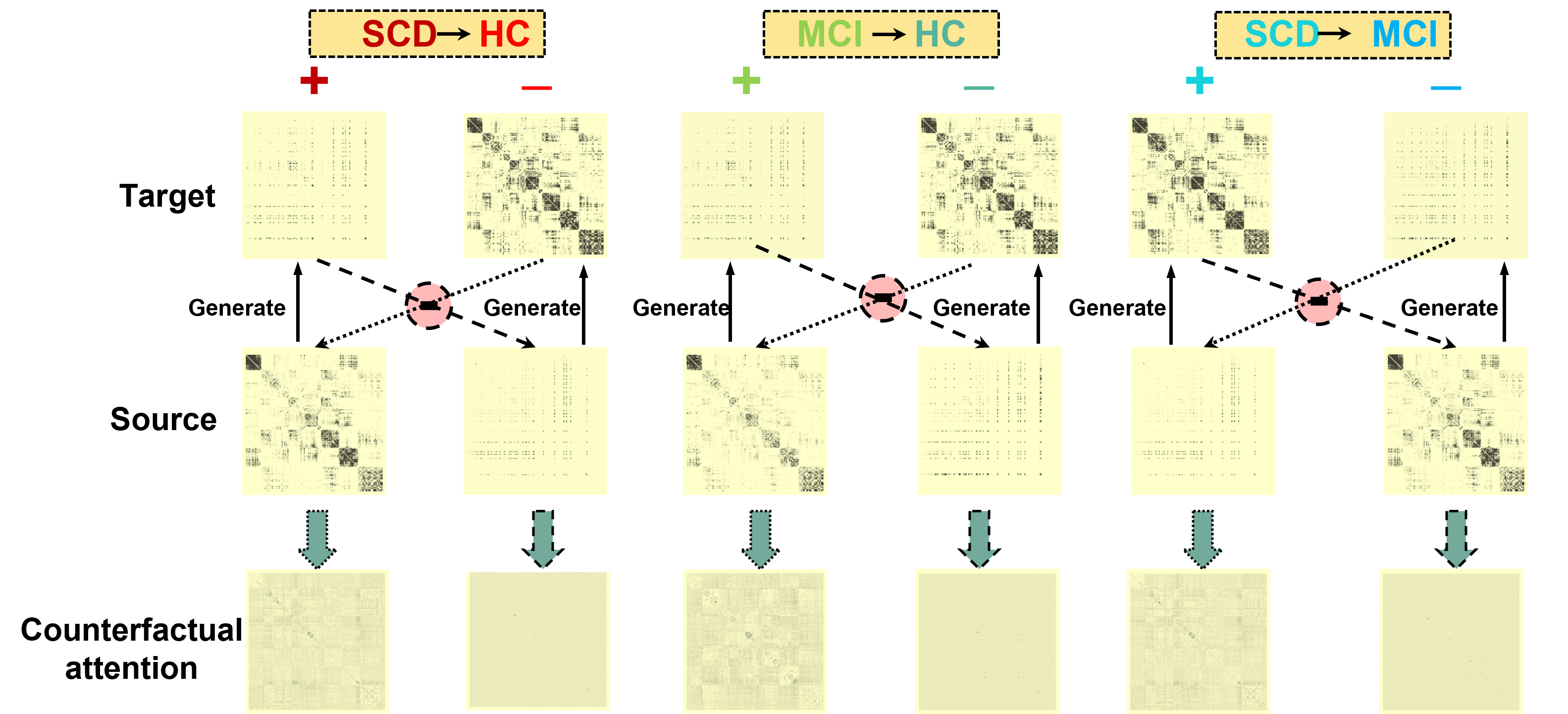}
\caption{Visualization of multimodal counterfactual attention maps across different cognitive-state transition tasks. The upper and lower panels show bidirectional transitions among HC, SCD, and MCI. For each transition, the source-label FC/SC matrices and generated target-label FC/SC matrices are compared to obtain positive and negative counterfactual attention maps.}
\label{fig:multimodal_attention}
\end{figure}

To further evaluate the interpretability of the multimodal GCAN, we visualize the counterfactual attention maps generated from joint FC--SC modeling, as shown in Fig.~\ref{fig:multimodal_attention}. For each cognitive-state transition, the source-label FC/SC matrices and the generated target-label FC/SC matrices are compared, and their differences are projected onto brain regions to obtain positive and negative counterfactual attention maps. The positive attention reflects connectivity patterns that become more prominent when the source state is transformed toward the target state, whereas the negative attention indicates weakened or suppressed connectivity patterns.

The multimodal counterfactual attention maps reveal complementary FC and SC alteration patterns across cognitive-state transitions. FC attention shows more dynamic network reorganization: responses are relatively localized in the HC-to-SCD transformation but already emerge around module boundaries, while HC-to-MCI and MCI-to-SCD transformations exhibit broader attention extending to cross-module connections, suggesting a progression from local functional reconfiguration to distributed network disruption. In contrast, SC attention is more stable and sparse, consistent with the relatively constrained nature of anatomical connectivity, and mainly highlights localized topology-related structural alterations in cognition-related networks. Jointly, FC appears more sensitive to early functional reorganization, whereas SC provides complementary evidence of anatomical disruption. Their partial overlap in high-order cognitive networks, including the DMN, FPN, and CON, supports the role of structure--function coupling abnormalities in cognitive decline and indicates that multimodal GCAN identifies sparse, biologically meaningful FC--SC changes rather than nonspecific whole-connectome responses.

\begin{figure}
\centering
\includegraphics[width=\textwidth]{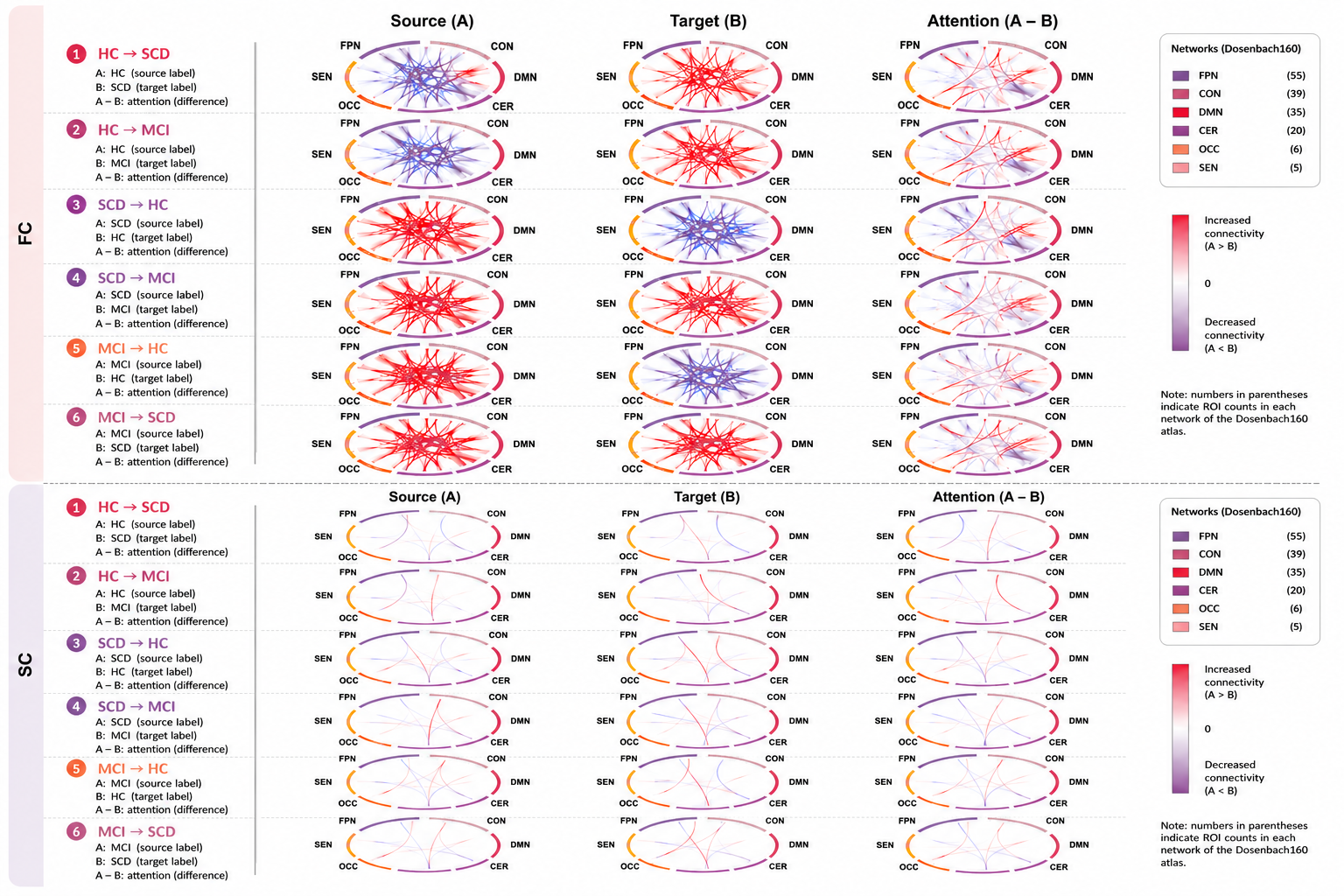}
\caption{Multimodal circular connectome visualization of FC and SC counterfactual analysis across different cognitive-state transition tasks. The upper panel shows functional connectivity (FC), and the lower panel shows structural connectivity (SC). For each modality, source-label connectomes, generated target-label connectomes, and counterfactual attention maps are shown from left to right. Red and purple/blue connections indicate increased and decreased connectivity, respectively. The outer ring represents the Dosenbach160 functional networks, including FPN, CON, DMN, CER, OCC, and SEN.}
\label{fig:dual_connectome}
\end{figure}

\subsubsection{Multimodal circular connectome analysis}
\label{sec:multimodal_circular}

To further analyze structure--function network reorganization, we draw circular connectome diagrams for both FC and SC, as shown in Fig.~\ref{fig:dual_connectome}. For each modality, the source-label connectome, generated target-label connectome, and counterfactual attention are shown in three columns. The upper part presents FC-based circular connectomes, while the lower part presents SC-based circular connectomes. This visualization provides a direct comparison between functional reorganization and structural topology changes during different cognitive-state transitions.

The multimodal circular connectome diagrams further show that FC and SC provide complementary explanations. FC attention captures widespread cross-network functional reorganization, especially involving the DMN, FPN, CON, and SEN. Compared with the source and target FC patterns, the counterfactual attention highlights a limited number of discriminative connections rather than the whole connectome, indicating that the proposed method can identify sparse disease-transition-related functional changes.

In contrast, SC attention is generally more stable and sparse. This is consistent with the biological property that structural connectivity reflects relatively stable anatomical pathways, whereas FC is more sensitive to dynamic functional synchronization. Although SC attention contains fewer highlighted connections than FC attention, it still reveals network-specific topology changes related to cognitive decline. These structural changes provide complementary evidence for the functional abnormalities observed in FC.

Across tasks, the attention patterns also show a gradual change from early subjective decline to mild impairment. In HC vs. SCD, the attention is relatively localized, suggesting subtle structure--function reorganization at the early stage of cognitive decline. In HC vs. MCI, the attention becomes more extensive, indicating that MCI involves more pronounced disruption in both functional interactions and structural pathways. In SCD vs. MCI, the discriminative connections are mainly concentrated in high-order cognitive networks, suggesting that fine-grained progression from subjective decline to mild impairment is associated with subtle but meaningful structure--function coordination changes.

Overall, the FC and SC circular connectome results support the effectiveness of multimodal counterfactual reasoning. FC provides sensitive evidence of functional reorganization, whereas SC provides topology-constrained evidence of anatomical disruption. Their complementary patterns indicate that cognitive decline is not reflected by a single modality alone, but by coordinated abnormalities in functional and structural brain networks.

\subsubsection{Statistical analysis of multimodal diagnostic performance}
\label{sec:multimodal_statistics}
\begin{figure}
\centering
\includegraphics[width=0.9\textwidth]{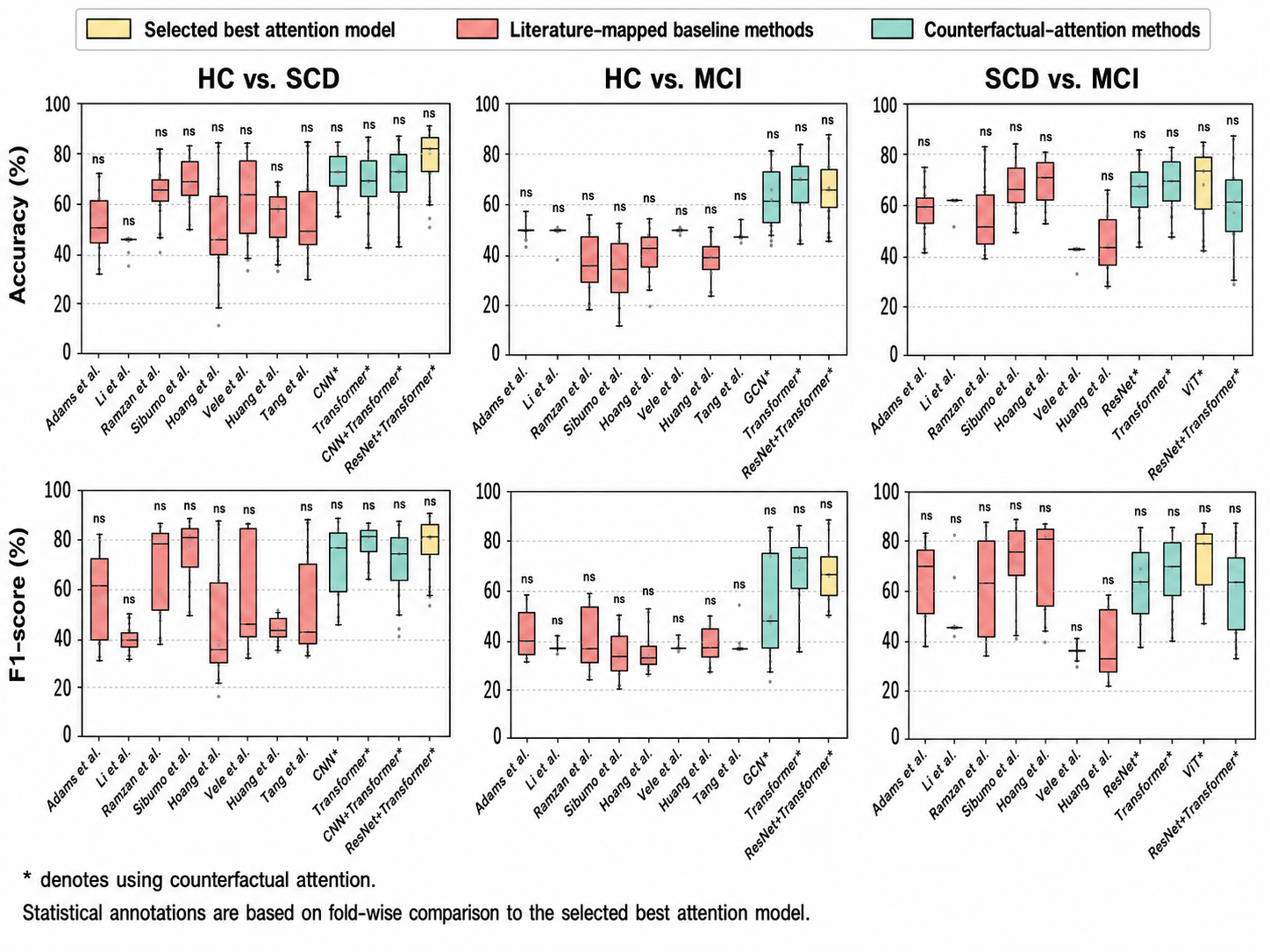}
\caption{Statistical comparison of multimodal diagnostic performance across three classification tasks. Boxplots show the fold-wise distributions of ACC, recall, precision, and F1-score. Yellow boxes indicate the selected best attention model, red boxes indicate literature-mapped baseline methods, and cyan boxes indicate counterfactual-attention-based methods. $^{*}$ denotes using counterfactual attention. Statistical annotations are based on fold-wise comparison with the selected best attention model.}
\label{fig:multimodal_statics}
\end{figure}

To further examine the performance distribution of different methods across five-fold cross-validation, we conduct a fold-wise statistical analysis based on ACC, recall, precision, and F1-score. As shown in Fig.~\ref{fig:multimodal_statics}, the boxplots compare literature-mapped baseline methods, counterfactual-attention-based methods, and the selected best attention model for each classification task. The statistical annotations are obtained by comparing each method with the selected best attention model across folds.

Overall, the boxplot distributions provide a descriptive view of the fold-wise performance variation rather than strong statistical evidence of method superiority. In the HC vs. SCD task, several counterfactual-attention-based models show relatively higher median ACC and F1-score than most literature-mapped baselines. This suggests that counterfactual attention may help capture early structure--function alterations related to SCD. However, the number of validation folds is limited, and several methods still show noticeable inter-fold variation. Therefore, this observation should be interpreted as a favorable trend rather than a statistically conclusive improvement.

For the HC vs. MCI task, the selected attention-based models also show competitive performance distributions in ACC and F1-score. Nevertheless, the performance spread across folds is relatively large for several methods, indicating sensitivity to sample partitioning and cohort heterogeneity. As a result, although the proposed counterfactual attention mechanism appears to provide useful discriminative information, the current results are insufficient to establish robust statistical superiority over all comparison methods.

For the SCD vs. MCI task, the distributions are more variable and the distinction among methods is less stable. Some baseline methods achieve competitive ACC or recall, while attention-based models show advantages only in certain metrics or folds. This is consistent with the fine-grained nature of SCD vs. MCI classification, where disease-related FC--SC differences are relatively subtle and the available paired multimodal samples remain limited.

Most pairwise comparisons do not reach statistical significance. This is expected because the analysis is based on five-fold cross-validation, resulting in a small number of paired observations for statistical testing. Therefore, the boxplots should be viewed mainly as a visualization of performance variability. To further describe this uncertainty, we report fold-level 95\% confidence intervals for ACC and F1-score in Table~\ref{tab:multitask_ci_acc_f1}. The confidence intervals are generally wide for several models, especially in more difficult tasks such as HC vs. MCI and SCD vs. MCI. These results indicate that the observed improvements should be interpreted cautiously. Overall, the statistical analysis suggests that multimodal counterfactual attention shows promising but preliminary trends for FC--SC-based cognitive decline diagnosis, and larger paired multimodal cohorts are needed for more reliable statistical validation.

\begin{table*}[htbp]
\centering
\scriptsize
\renewcommand{\arraystretch}{1.10}
\setlength{\tabcolsep}{2.2pt}

\caption{Fold-level 95\% confidence intervals of ACC and F1-score across three classification tasks. $^{*}$ denotes using counterfactual attention.}
\label{tab:multitask_ci_acc_f1}

\resizebox{\textwidth}{!}{
\begin{tabular}{lcccccc}
\toprule
\multirow{2}{*}{\textbf{Method}} 
& \multicolumn{2}{c}{\textbf{HC vs. SCD}} 
& \multicolumn{2}{c}{\textbf{HC vs. MCI}} 
& \multicolumn{2}{c}{\textbf{SCD vs. MCI}} \\
\cmidrule(lr){2-3} \cmidrule(lr){4-5} \cmidrule(lr){6-7}
& \textbf{ACC 95\% CI} & \textbf{F1 95\% CI}
& \textbf{ACC 95\% CI} & \textbf{F1 95\% CI}
& \textbf{ACC 95\% CI} & \textbf{F1 95\% CI} \\
\midrule

Adarsh et al.~\cite{adarsh2024multimodal} 
& [34.54, 66.80] & [21.70, 61.40]
& [37.41, 55.93] & [24.45, 52.21]
& [39.58, 76.42] & [21.98, 71.60] \\

Li et al.~\cite{li2023identification} 
& [34.96, 42.38] & [25.87, 29.85]
& [12.24, 67.76] & [8.16, 45.18]
& [43.54, 74.46] & [21.21, 64.55] \\

Ramzan et al.~\cite{ramzan2020deep} 
& [43.38, 79.28] & [28.70, 78.20]
& [6.10, 93.90] & [0.00, 88.67]
& [33.22, 74.78] & [21.80, 74.30] \\

Sibilano et al.~\cite{sibilano2024understanding} 
& [38.82, 86.52] & [32.85, 84.77]
& [0.00, 87.07] & [0.00, 82.37]
& [54.89, 79.11] & [41.61, 79.87] \\

Zuo et al.~\cite{zuo2023alzheimer} 
& [3.20, 84.80] & [0.00, 69.01]
& [5.90, 57.44] & [2.51, 55.49]
& [54.41, 95.59] & [31.28, 98.52] \\

Feng et al.~\cite{feng2025crossmodal} 
& [31.60, 85.74] & [15.98, 84.44]
& [27.04, 56.30] & [18.17, 46.49]
& [19.40, 70.60] & [8.37, 64.39] \\

Huang et al.~\cite{huang2026transformer} 
& [31.18, 62.16] & [24.31, 38.55]
& [12.24, 67.76] & [8.16, 45.18]
& [34.89, 57.11] & [27.19, 40.01] \\

Tang et al.~\cite{tang2024multimodal} 
& [26.82, 74.52] & [14.07, 74.13]
& [27.04, 56.30] & [18.17, 46.49]
& [13.96, 76.04] & [7.56, 72.34] \\

\midrule

CNN$^{*}$ 
& [61.63, 85.04] & [35.56, 88.40]
& [33.16, 70.18] & [23.27, 69.40]
& [25.22, 66.78] & [15.94, 53.82] \\

ResNet$^{*}$ 
& [54.54, 76.13] & [24.56, 66.92]
& [22.85, 70.49] & [13.41, 67.25]
& [39.79, 84.21] & [26.49, 80.89] \\

GCN$^{*}$ 
& [30.11, 55.22] & [23.86, 35.42]
& [44.07, 62.59] & [30.96, 38.38]
& [42.89, 65.11] & [34.19, 40.15] \\

Transformer$^{*}$ 
& [41.15, 92.18] & [34.72, 90.05]
& [12.93, 100.00] & [7.16, 100.00]
& [43.56, 82.44] & [34.63, 79.51] \\

CNN+Transformer$^{*}$ 
& [39.36, 100.00] & [24.85, 100.00]
& [11.00, 79.00] & [7.63, 75.03]
& [34.89, 57.11] & [27.18, 40.01] \\

GCN+Transformer$^{*}$ 
& [31.18, 62.15] & [24.31, 38.54]
& [37.41, 55.92] & [27.04, 36.29]
& [37.58, 62.42] & [27.55, 38.64] \\

ResNet+Transformer$^{*}$ 
& [41.51, 97.16] & [33.57, 98.33]
& [33.16, 70.18] & [19.12, 66.88]
& [36.38, 73.62] & [20.61, 69.91] \\

\bottomrule
\end{tabular}
}

\vspace{1mm}
\begin{flushleft}
\footnotesize
\textit{Note}: Only ACC and F1-score confidence intervals are reported to avoid repeating the complete mean performance values already shown in Table~\ref{tab:multitask_compare_merge}. The 95\% confidence intervals are calculated from five-fold results using Student's $t$ distribution and clipped to the valid range of [0, 100]. F1-score denotes macro-F1, consistent with the main performance table. $^{*}$ denotes models using counterfactual attention. These intervals are intended to visualize cross-validation uncertainty rather than to establish statistical superiority.
\end{flushleft}
\end{table*}

\subsubsection{Multimodal FC--SC synthesis quality analysis}
\label{sec:multimodal_similarity}

To further evaluate the synthesis quality of multimodal GCAN during counterfactual FC--SC generation, we assess the generated connectomes from complementary matrix-level and edge-level perspectives. Since FC and SC characterize different aspects of the brain connectome, a reliable multimodal generator should preserve not only the global matrix structure, but also the relative connectivity organization and ROI-to-ROI connection strength in both modalities. Therefore, PSNR, SSIM, Pearson correlation coefficient, MAE, and MSE are jointly used for evaluation. Specifically, PSNR measures global intensity-level consistency, SSIM evaluates local structural similarity, and Pearson correlation measures whether the generated connectome preserves the relative connectivity pattern of the reference matrix. In addition, MAE and MSE are computed after normalizing all generated and reference matrices to the same value range, providing direct measurements of numerical deviations at the connectivity-edge level. Higher PSNR, SSIM, and correlation values, together with lower MAE and MSE, indicate better FC--SC synthesis quality. The results are summarized in Table~\ref{tab:fc_sc_synthesis_quality}.

\begin{table}[htbp]
\centering
\scriptsize
\renewcommand{\arraystretch}{1.12}
\setlength{\tabcolsep}{2.2pt}

\caption{Matrix-level similarity, edge-wise reconstruction errors, and correlation coefficients of multimodal FC--SC synthesis under different classification tasks and counterfactual branches.}
\label{tab:fc_sc_synthesis_quality}

\resizebox{\textwidth}{!}{
\begin{tabular}{llccccccccccccccc}
\toprule
\multirow{2}{*}{\textbf{Task}} 
& \multirow{2}{*}{\textbf{Branch}} 
& \multicolumn{5}{c}{\textbf{FC}} 
& \multicolumn{5}{c}{\textbf{SC}} 
& \multicolumn{5}{c}{\textbf{Average}} \\
\cmidrule(lr){3-7} \cmidrule(lr){8-12} \cmidrule(lr){13-17}
& & \textbf{PSNR} & \textbf{SSIM} & \textbf{MAE} & \textbf{MSE} & \textbf{Corr.}
  & \textbf{PSNR} & \textbf{SSIM} & \textbf{MAE} & \textbf{MSE} & \textbf{Corr.}
  & \textbf{PSNR} & \textbf{SSIM} & \textbf{MAE} & \textbf{MSE} & \textbf{Corr.} \\
\midrule

\multirow{2}{*}{HC vs. SCD}
& Positive 
& 28.1761 & 0.6991 & 0.0128 & 0.0017 & 0.3623
& 18.4525 & 0.5832 & 0.0634 & 0.0154 & 0.7943
& 23.3143 & 0.6412 & 0.0381 & 0.0085 & 0.5783 \\
& Negative 
& 32.3128 & 0.8700 & 0.0064 & 0.0007 & 0.7349
& 19.0175 & 0.5182 & 0.0605 & 0.0136 & 0.7469
& 25.6652 & 0.6941 & 0.0334 & 0.0072 & 0.7409 \\

\midrule

\multirow{2}{*}{HC vs. MCI}
& Positive 
& 31.0282 & 0.8630 & 0.0096 & 0.0011 & 0.7578
& 16.9546 & 0.4667 & 0.0725 & 0.0213 & 0.6551
& 23.9914 & 0.6648 & 0.0411 & 0.0112 & 0.7064 \\
& Negative 
& 28.9345 & 0.7924 & 0.0233 & 0.0016 & 0.6474
& 17.9061 & 0.4458 & 0.0676 & 0.0174 & 0.7041
& 23.4203 & 0.6191 & 0.0455 & 0.0095 & 0.6758 \\

\midrule

\multirow{2}{*}{MCI vs. SCD}
& Positive 
& 19.4013 & 0.6132 & 0.0579 & 0.0126 & 0.8259
& 26.9447 & 0.6451 & 0.0144 & 0.0023 & 0.3201
& 23.1730 & 0.6292 & 0.0362 & 0.0074 & 0.5730 \\
& Negative 
& 18.4084 & 0.5396 & 0.0621 & 0.0154 & 0.7189
& 28.1181 & 0.7885 & 0.0240 & 0.0019 & 0.5876
& 23.2632 & 0.6640 & 0.0430 & 0.0086 & 0.6532 \\

\midrule

\multirow{3}{*}{Average}
& Positive 
& 26.2019 & 0.7251 & 0.0268 & 0.0051 & 0.6487
& 20.7839 & 0.5650 & 0.0501 & 0.0130 & 0.5898
& 23.4929 & 0.6450 & 0.0384 & 0.0091 & 0.6192 \\
& Negative 
& 26.5519 & 0.7340 & 0.0306 & 0.0059 & 0.7004
& 21.6806 & 0.5842 & 0.0507 & 0.0110 & 0.6795
& 24.1162 & 0.6591 & 0.0406 & 0.0084 & 0.6900 \\
& All 
& 26.3769 & 0.7296 & 0.0287 & 0.0055 & 0.6745
& 21.2322 & 0.5746 & 0.0504 & 0.0120 & 0.6347
& 23.8046 & 0.6521 & 0.0395 & 0.0088 & 0.6546 \\

\bottomrule
\end{tabular}
}

\end{table}

As shown in Table~\ref{tab:fc_sc_synthesis_quality}, multimodal GCAN achieves an average PSNR of $23.8046$ dB, SSIM of $0.6521$, Pearson correlation coefficient of $0.6546$, normalized MAE of $0.0395$, and normalized MSE of $0.0088$ across all tasks, branches, and modalities. These results indicate that the generated FC--SC connectomes preserve a reasonable degree of global matrix structure, relative connectivity organization, and edge-level numerical fidelity. Compared with SC, FC shows higher average PSNR, SSIM, and correlation, as well as lower MAE and MSE, suggesting that FC synthesis is numerically closer to the reference connectomes in the current setting. SC synthesis is relatively more challenging, possibly because SC has sparser topology and stronger anatomical constraints, making edge-wise reconstruction more sensitive to structural variability.

The branch-wise results show that positive and negative branches exhibit comparable but not identical synthesis behavior. The negative branch obtains slightly higher average PSNR, SSIM, and Pearson correlation, whereas the positive branch shows slightly lower average MSE. This indicates that the two counterfactual directions preserve target-state connectivity information from different perspectives, rather than showing a consistent one-sided advantage. Across tasks, HC vs. SCD and HC vs. MCI show relatively stable multimodal synthesis quality, while MCI vs. SCD exhibits more heterogeneous modality-specific behavior. For example, the MCI vs. SCD positive branch shows high FC correlation but lower SC correlation, suggesting that functional and structural connectomes may differ in their reconstruction difficulty for fine-grained cognitive-state transitions.

Overall, these metrics provide complementary evidence for evaluating multimodal synthesis quality. PSNR reflects whether the generated connectomes preserve the global magnitude distribution, SSIM evaluates the retention of local matrix structures, Pearson correlation measures the consistency of relative connectivity organization, and MAE/MSE quantify absolute deviations of individual ROI-to-ROI edges after normalization. The agreement among these indicators suggests that multimodal GCAN can generate FC--SC connectomes with reasonable structural and numerical fidelity. This provides quantitative support for the reliability of the generated multimodal connectomes used in subsequent counterfactual attention analysis, while the results should still be interpreted as synthesis-quality evidence rather than direct proof of biological causality.

\subsubsection{Structure--function counterfactual overlap analysis}
\label{sec:sf_overlap}

\begin{table}[htbp]
\centering
\footnotesize
\renewcommand{\arraystretch}{1.15}
\setlength{\tabcolsep}{5pt}

\caption{Top-$k$ overlap between FC and SC counterfactual attention edges across six source-to-target cognitive-state transitions. The overlap is computed using the Jaccard index between selected top-$k\%$ FC and SC attention edges.}
\label{tab:topk_fc_sc_overlap_hc_mci}

\begin{tabular*}{\textwidth}{@{\extracolsep{\fill}}lcccccc@{}}
\toprule
\multirow{2}{*}{\textbf{Transition}} 
& \multicolumn{3}{c}{\textbf{Positive attention overlap}} 
& \multicolumn{3}{c}{\textbf{Negative attention overlap}} \\
\cmidrule(lr){2-4} \cmidrule(lr){5-7}
& \textbf{Top-5\%} & \textbf{Top-10\%} & \textbf{Top-20\%}
& \textbf{Top-5\%} & \textbf{Top-10\%} & \textbf{Top-20\%} \\
\midrule
HC $\rightarrow$ MCI & 2.25\% & 5.82\% & 10.75\% & 2.00\% & 4.82\% & 10.83\% \\
HC $\rightarrow$ SCD & 2.25\% & 5.25\% & 11.09\% & 2.17\% & 4.48\% & 10.49\%  \\
SCD $\rightarrow$ MCI & 1.76\% & 5.17\% & 9.14\% & 6.53\% & 8.12\% & 12.84\%  \\
MCI $\rightarrow$ HC & 2.50\% & 5.21\% & 11.09\% & 2.17\% & 4.73\% & 9.14\%  \\
MCI $\rightarrow$ SCD & 3.00\% & 6.35\% & 10.73\% & 1.76\% & 3.67\% & 9.66\%  \\
SCD $\rightarrow$ HC & 2.75\% & 5.17\% & 11.41\% & 5.21\% & 8.81\% & 11.97\% \\
\bottomrule
\end{tabular*}

\vspace{1mm}
\end{table}

To further examine whether the FC and SC counterfactual attention maps identify shared disease-related connections, we compute the top-$k$ overlap between the two modalities. For each source-to-target transition, the upper triangular elements of the FC and SC counterfactual attention matrices are used to avoid duplicated symmetric connections. We separately calculate the overlap for positive and negative attention edges using the Jaccard index. Positive overlap reflects shared FC--SC connections that become more prominent during source-to-target transformation, whereas negative overlap reflects shared connections that are suppressed.

As shown in Table~\ref{tab:topk_fc_sc_overlap_hc_mci}, the FC--SC overlap values are generally modest at the top-5\% and top-10\% levels, but become more evident at the top-20\% level. Across the six transition directions, the positive top-20\% overlap ranges from 9.14\% to 11.41\%, while the negative top-20\% overlap ranges from 9.14\% to 12.84\%. These results indicate that FC and SC counterfactual attention maps do not simply highlight identical connections. Instead, they provide complementary explanations with a stable subset of shared disease-transition-related edges.

A more detailed comparison further shows that the overlap pattern depends on the transition direction. For HC $\rightarrow$ SCD and HC $\rightarrow$ MCI, both positive and negative overlaps remain relatively balanced, suggesting that early cognitive-state transformation is mainly characterized by partially shared but still modality-specific functional and structural changes. In contrast, the SCD $\rightarrow$ MCI and SCD $\rightarrow$ HC transitions show higher negative overlap, reaching 12.84\% and 11.97\% at the top-20\% level, respectively. This suggests that FC and SC tend to agree more on suppressed connections in these transitions, which may reflect shared structure--function alterations associated with the progression or reversal of cognitive decline patterns.

Overall, the top-$k$ overlap analysis provides quantitative support for partial structure--function coordination in the counterfactual attention space. The results also suggest that FC and SC are not redundant: FC attention captures more widespread functional reorganization, whereas SC attention provides sparse structural evidence. Their overlap among top-ranked attention edges complements the multimodal circular connectome visualization and supports the use of joint FC--SC counterfactual reasoning for interpretable cognitive decline diagnosis.

\subsubsection{Multimodal ablation study}
\label{sec:multimodal_ablation}

\textbf{Effect of counterfactual attention on multimodal diagnosis.}
As shown in Table~\ref{tab:multimodal_attention_ablation}, introducing counterfactual attention generally improves multimodal diagnostic performance across the three classification tasks, although the magnitude of improvement varies across backbones and tasks. In HC vs. SCD, the improvement is relatively clear. For example, CNN accuracy increases from $50.67\%$ to $73.33\%$, and F1-score increases from $41.55\%$ to $61.98\%$. ViT also shows improved ACC and F1-score after introducing counterfactual attention. These results suggest that counterfactual attention may help multimodal models capture early structure--function alterations related to SCD. In HC vs. MCI, attention-based variants also show performance gains for GCN, Transformer, and ViT, indicating that counterfactual attention provides useful auxiliary information for distinguishing cognitively impaired subjects from healthy controls. In SCD vs. MCI, the improvement is more moderate and less uniform, which is consistent with the fine-grained nature of this task. Overall, the diagnostic ablation results support the usefulness of counterfactual attention, while also indicating that the benefit depends on the backbone architecture and task difficulty.

\begin{table}[htbp]
\centering
\footnotesize
\setlength{\tabcolsep}{2pt}
\renewcommand{\arraystretch}{1.08}

\caption{Ablation study of counterfactual attention across different multimodal classification tasks (\%, five-fold cross-validation). $^{*}$ denotes using counterfactual attention.}
\label{tab:multimodal_attention_ablation}

\begin{tabular*}{\textwidth}{@{\extracolsep{\fill}}llccccc@{}}
\toprule
\multirow{2}{*}{\textbf{Task}} 
& \multirow{2}{*}{\textbf{Method}} 
& \multirow{2}{*}{\textbf{CA}} 
& \multicolumn{4}{c}{\textbf{Classification performance}} \\
\cmidrule(lr){4-7}
& & & \textbf{ACC} & \textbf{Recall} & \textbf{Precision} & \textbf{F1} \\
\midrule

\multirow{6}{*}{HC vs. SCD}
& CNN & \ding{55} & $50.67 \pm 11.62$ & $46.67 \pm 15.46$ & $42.67 \pm 20.37$ & $41.55 \pm 14.30$ \\
& CNN$^{*}$ & \ding{52} & $\mathbf{73.33 \pm 8.43}$ & $\mathbf{66.67 \pm 13.94}$ & $\mathbf{64.33 \pm 26.74}$ & $\mathbf{61.98 \pm 19.03}$ \\
& ResNet & \ding{55} & $61.33 \pm 12.93$ & $\mathbf{60.00 \pm 16.16}$ & $\mathbf{58.50 \pm 25.77}$ & $\mathbf{53.45 \pm 17.83}$ \\
& ResNet$^{*}$ & \ding{52} & $\mathbf{65.33 \pm 7.77}$ & $56.67 \pm 9.72$ & $47.83 \pm 22.49$ & $49.90 \pm 15.28$ \\
& ViT & \ding{55} & $44.00 \pm 29.39$ & $40.00 \pm 25.50$ & $32.00 \pm 29.97$ & $33.57 \pm 25.53$ \\
& ViT$^{*}$ & \ding{52} & $\mathbf{61.33 \pm 12.93}$ & $\mathbf{58.33 \pm 15.81}$ & $\mathbf{55.17 \pm 24.46}$ & $\mathbf{53.45 \pm 17.83}$ \\

\midrule
\multirow{6}{*}{HC vs. MCI}
& GCN & \ding{55} & $40.00 \pm 20.00$ & $40.00 \pm 20.00$ & $20.00 \pm 10.00$ & $26.67 \pm 13.33$ \\
& GCN$^{*}$ & \ding{52} & $\mathbf{53.33 \pm 6.67}$ & $\mathbf{50.00 \pm 0.00}$ & $\mathbf{26.67 \pm 3.33}$ & $\mathbf{34.67 \pm 2.67}$ \\
& Transformer & \ding{55} & $40.00 \pm 33.91$ & $40.00 \pm 33.91$ & $31.67 \pm 35.12$ & $34.67 \pm 34.36$ \\
& Transformer$^{*}$ & \ding{52} & $\mathbf{60.00 \pm 33.91}$ & $\mathbf{60.00 \pm 33.91}$ & $\mathbf{58.33 \pm 38.73}$ & $\mathbf{56.00 \pm 35.18}$ \\
& ViT & \ding{55} & $31.67 \pm 18.56$ & $30.00 \pm 18.71$ & $28.33 \pm 19.44$ & $29.00 \pm 19.08$ \\
& ViT$^{*}$ & \ding{52} & $\mathbf{60.00 \pm 33.91}$ & $\mathbf{60.00 \pm 33.91}$ & $\mathbf{63.33 \pm 35.59}$ & $\mathbf{59.33 \pm 33.63}$ \\

\midrule
\multirow{6}{*}{SCD vs. MCI}
& ResNet & \ding{55} & $54.00 \pm 14.97$ & $55.00 \pm 17.95$ & $50.00 \pm 24.72$ & $48.05 \pm 18.91$ \\
& ResNet$^{*}$ & \ding{52} & $\mathbf{63.00 \pm 14.00}$ & $\mathbf{58.33 \pm 15.81}$ & $\mathbf{56.83 \pm 25.99}$ & $\mathbf{54.79 \pm 18.98}$ \\
& GCN+Transformer & \ding{55} & $46.00 \pm 8.00$ & $48.33 \pm 3.33$ & $\mathbf{27.33 \pm 8.07}$ & $\mathbf{33.60 \pm 4.62}$ \\
& GCN+Transformer$^{*}$ & \ding{52} & $\mathbf{50.00 \pm 8.94}$ & $\mathbf{50.00 \pm 0.00}$ & $25.00 \pm 4.47$ & $33.10 \pm 3.99$ \\
& ResNet+Transformer & \ding{55} & $45.00 \pm 22.36$ & $41.67 \pm 21.73$ & $40.83 \pm 27.94$ & $39.95 \pm 23.33$ \\
& ResNet+Transformer$^{*}$ & \ding{52} & $\mathbf{55.00 \pm 13.42}$ & $\mathbf{50.00 \pm 15.81}$ & $\mathbf{44.33 \pm 23.11}$ & $\mathbf{45.26 \pm 17.76}$ \\

\bottomrule
\end{tabular*}

\vspace{1mm}
\begin{flushleft}
\footnotesize
\textit{Note}: CA denotes counterfactual attention. \ding{52} indicates that counterfactual attention is used, and \ding{55} indicates that it is not used.
\end{flushleft}
\end{table}

\textbf{Synthesis quality of generated multimodal connectomes.}
In addition to diagnostic performance, we further evaluate whether the generated target-label FC and SC matrices preserve meaningful connectome structures. As shown in Table~\ref{tab:fc_sc_synthesis_quality}, PSNR and SSIM characterize matrix-level similarity, Pearson correlation measures the preservation of relative connectivity organization, and MAE/MSE quantify edge-wise numerical deviations after normalization. These metrics provide complementary evidence for assessing the generated counterfactual connectomes. The results show that the generated FC and SC matrices retain a certain degree of matrix-level and edge-level fidelity across different transition directions. However, the synthesis quality differs between FC and SC. FC reflects dynamic functional reorganization and therefore tends to show larger variability, whereas SC is relatively sparse and topologically stable. Thus, these synthesis-quality metrics should be interpreted as evidence that the generated connectomes are numerically and structurally reasonable, rather than as direct proof of biological validity.

\begin{table*}[htbp]
\centering
\scriptsize
\renewcommand{\arraystretch}{1.08}
\setlength{\tabcolsep}{3pt}

\caption{Ablation study of AABT for multimodal FC--SC synthesis quality under different transition directions.}
\label{tab:aabt_synthesis_ablation_direction}

\begin{tabular*}{\textwidth}{@{\extracolsep{\fill}}lllcccccc@{}}
\toprule
\multirow{2}{*}{\textbf{Task}} 
& \multirow{2}{*}{\textbf{Transition}} 
& \multirow{2}{*}{\textbf{Modality}}
& \multicolumn{3}{c}{\textbf{w/o AABT}} 
& \multicolumn{3}{c}{\textbf{GCAN with AABT}} \\
\cmidrule(lr){4-6} \cmidrule(lr){7-9}
& & 
& \textbf{PSNR} & \textbf{SSIM} & \textbf{Corr.}
& \textbf{PSNR} & \textbf{SSIM} & \textbf{Corr.} \\
\midrule

\multirow{4}{*}{HC vs. SCD}
& \multirow{2}{*}{HC $\rightarrow$ SCD}
& FC & 19.6413 & 0.7655 & 0.8504 & 28.1761 & 0.6991 & 0.3623 \\
& & SC & 26.5783 & 0.9355 & 0.6552 & 18.4525 & 0.5832 & 0.7943 \\

& \multirow{2}{*}{SCD $\rightarrow$ HC}
& FC & 20.2961 & 0.7850 & 0.8700 & 32.3128 & 0.8700 & 0.7349 \\
& & SC & 26.2154 & 0.9340 & 0.5648 & 19.0175 & 0.5182 & 0.7469 \\

\midrule

\multirow{4}{*}{HC vs. MCI}
& \multirow{2}{*}{HC $\rightarrow$ MCI}
& FC & 19.7996 & 0.7572 & 0.8493 & 31.0282 & 0.8630 & 0.7578 \\
& & SC & 28.6956 & 0.9546 & 0.7098 & 16.9546 & 0.4667 & 0.6551 \\

& \multirow{2}{*}{MCI $\rightarrow$ HC}
& FC & 19.3784 & 0.7350 & 0.8340 & 28.9345 & 0.7924 & 0.6474 \\
& & SC & 28.5646 & 0.9520 & 0.7198 & 17.9061 & 0.4458 & 0.7041 \\

\midrule

\multirow{4}{*}{MCI vs. SCD}
& \multirow{2}{*}{MCI $\rightarrow$ SCD}
& FC & 19.4613 & 0.7535 & 0.8473 & 19.4013 & 0.6132 & 0.8259 \\
& & SC & 26.5104 & 0.9348 & 0.6439 & 26.9447 & 0.6451 & 0.3201 \\

& \multirow{2}{*}{SCD $\rightarrow$ MCI}
& FC & 20.1450 & 0.7792 & 0.8634 & 18.4084 & 0.5396 & 0.7189 \\
& & SC & 26.0749 & 0.9307 & 0.5825 & 28.1181 & 0.7885 & 0.5876 \\

\midrule

\multirow{3}{*}{Average}
& \multirow{3}{*}{All transitions}
& FC      & 19.7870 & 0.7626 & 0.8524 & 26.3769 & 0.7296 & 0.6745 \\
& & SC      & 27.1065 & 0.9403 & 0.6460 & 21.2322 & 0.5746 & 0.6347 \\
& & Average & 23.4467 & 0.8514 & 0.7492 & 23.8046 & 0.6521 & 0.6546 \\

\bottomrule
\end{tabular*}

\end{table*}

\begin{figure}
\centering
\includegraphics[width=\textwidth]{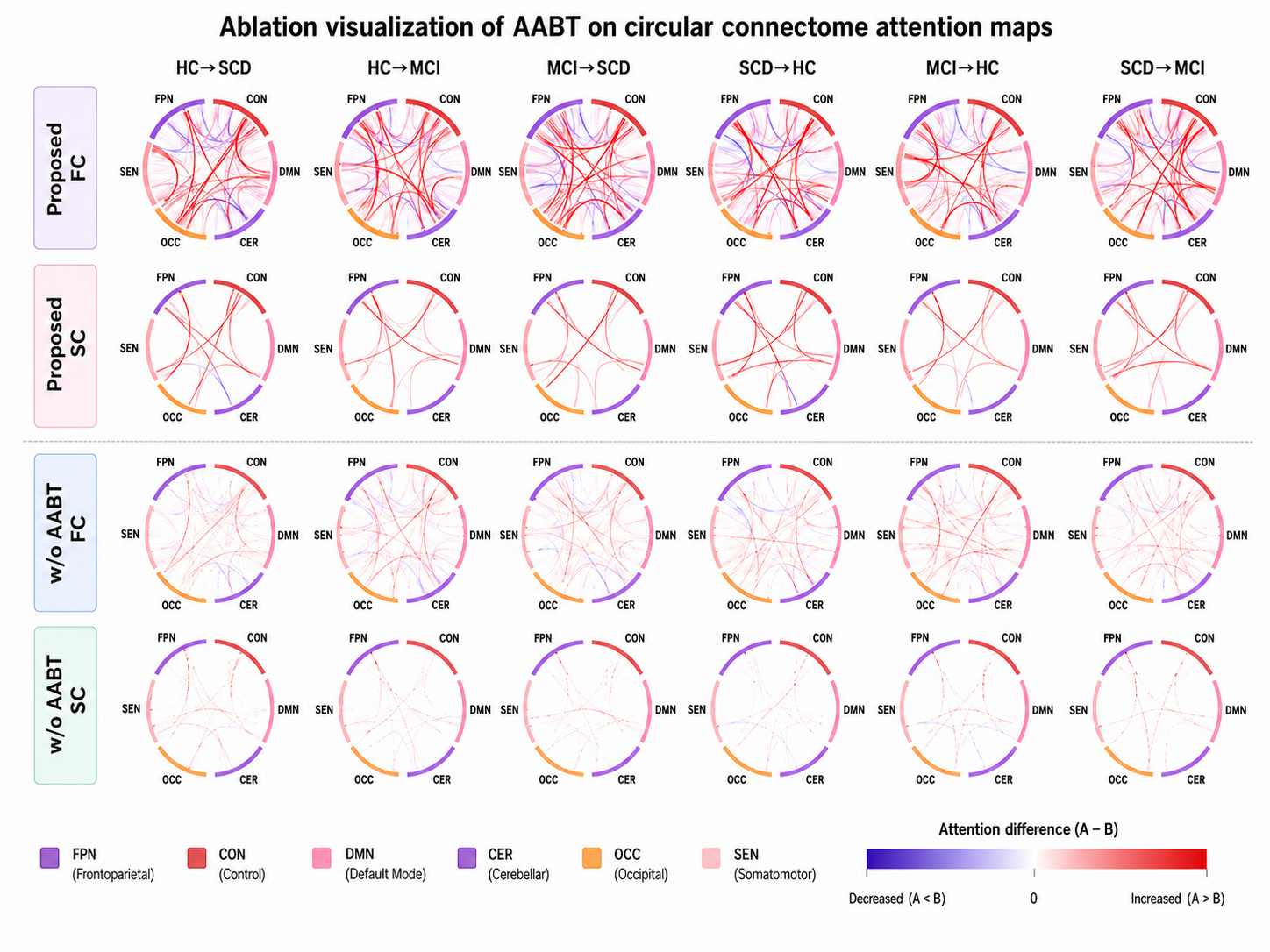}
\caption{Ablation visualization of AABT on multimodal circular connectome attention maps. The six columns correspond to six source-to-target transition directions. The four rows show FC attention with AABT, FC attention without AABT, SC attention with AABT, and SC attention without AABT, respectively. The visualization compares whether atlas-aware token encoding and decoding produce more structured network-level counterfactual attention.}
\label{fig:aabt_ablation_vis}
\end{figure}

\textbf{AABT ablation and atlas-aware topology preservation.}
We further examine the contribution of the Atlas-aware Bidirectional Transformer (AABT) by comparing the full GCAN model with a variant that removes atlas-aware network token encoding and decoding. The quantitative results are summarized in Table~\ref{tab:aabt_synthesis_ablation_direction}, and the corresponding visualization is shown in Fig.~\ref{fig:aabt_ablation_vis}. Instead of treating AABT as a component that should uniformly improve all reconstruction metrics, this ablation focuses on whether atlas-aware tokenization helps preserve structured connectome organization during counterfactual generation.

As shown in Table~\ref{tab:aabt_synthesis_ablation_direction}, the effect of AABT varies across modalities and transition directions. For FC synthesis, GCAN with AABT achieves higher PSNR in several transitions, including HC $\rightarrow$ SCD, SCD $\rightarrow$ HC, HC $\rightarrow$ MCI, and MCI $\rightarrow$ HC, indicating improved global reconstruction consistency for functional connectomes in these directions. For SC synthesis, the variant without AABT obtains higher SSIM in several cases, which may be related to the sparse and relatively stable topology of structural connectivity matrices. The full model achieves a slightly higher average PSNR, whereas the w/o AABT variant shows higher average SSIM and Pearson correlation. Therefore, the quantitative results suggest that AABT does not simply maximize matrix-level similarity, but changes the way the generator organizes FC--SC patterns under atlas constraints.

The visualization in Fig.~\ref{fig:aabt_ablation_vis} provides complementary evidence. Without AABT, the attention maps tend to be smoother or less organized at the network level, and the circular connectome patterns show weaker cross-network structure. After introducing AABT, the FC and SC attention maps present more concentrated and network-specific alterations across the six transition directions. These changes are mainly distributed among cognitive-related large-scale networks, such as DMN, FPN, CON, and SEN. This suggests that AABT mainly contributes to atlas-aware topology preservation and interpretable network-level organization, rather than to uniformly improving all low-level reconstruction scores. Together with the diagnostic ablation results, these findings support the role of AABT in producing more structured multimodal counterfactual attention.

\subsection{Comparison with CAM-based interpretation methods}
\label{sec:cam_comparison}

\begin{figure}
\centering
\includegraphics[width=0.9\textwidth]{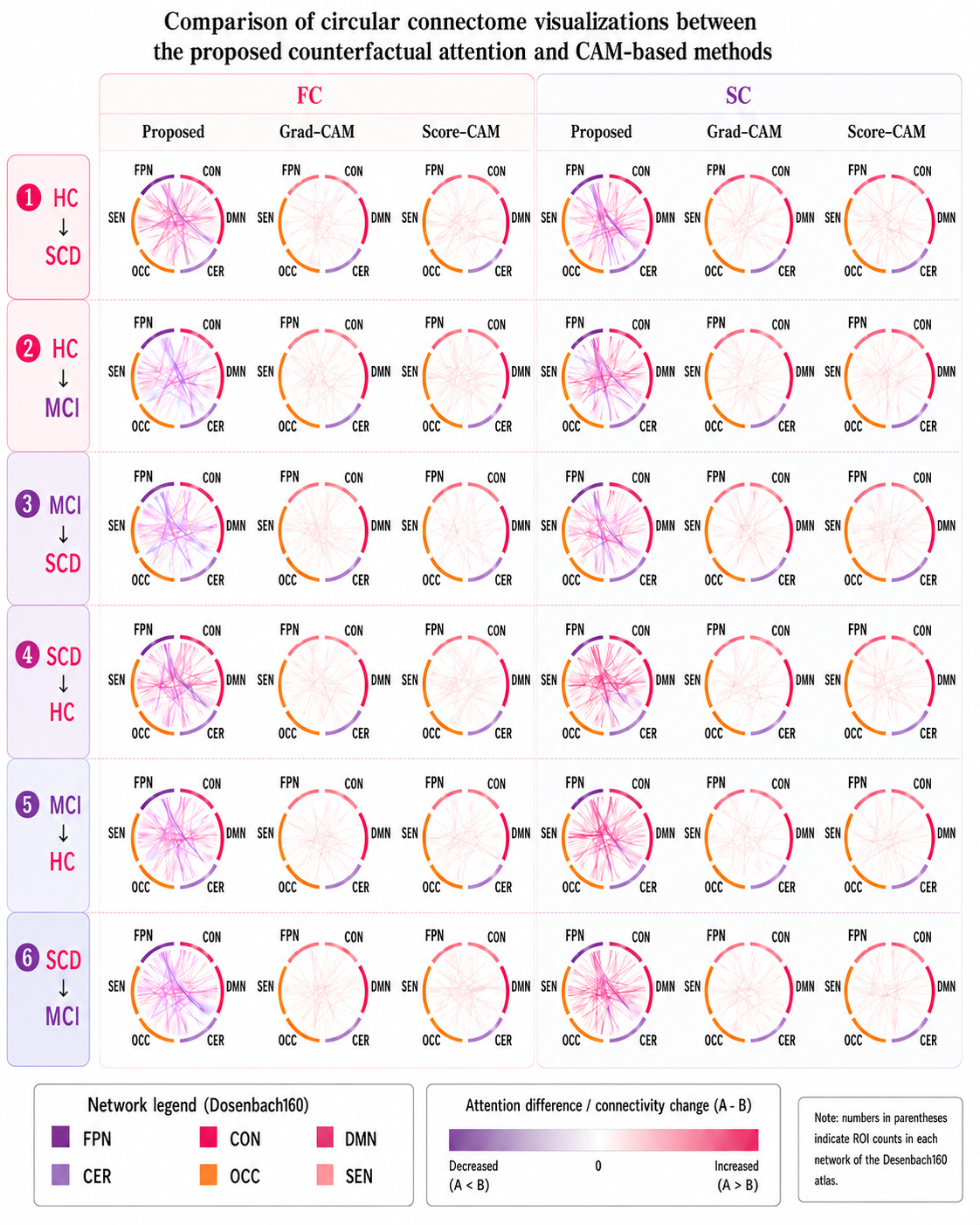}
\vspace{-5mm}
\caption{Comparison of circular connectome visualizations between the proposed counterfactual attention and CAM-based interpretation methods. Rows denote six cognitive-state transitions, and columns denote FC and SC modalities with the proposed method, Grad-CAM, and Score-CAM. The proposed method highlights more transition-specific and signed connectivity changes, whereas CAM-based methods mainly provide classifier-driven discriminative activation patterns.}
\label{fig:cam_connectome_comparison}
\vspace{-5mm}
\end{figure}

To further examine the interpretability of the proposed counterfactual attention mechanism, we compare it with two representative CAM-based explanation methods, Grad-CAM and Score-CAM. As shown in Fig.~\ref{fig:cam_connectome_comparison}, circular connectome visualizations are generated for six cognitive-state transitions across FC and SC modalities. For each modality, the proposed counterfactual attention map is compared with Grad-CAM and Score-CAM under the same transition setting.

Overall, the proposed method produces more structured and transition-specific connectivity patterns than the CAM-based methods. In the FC modality, the proposed counterfactual attention highlights distributed connections across FPN, DMN, CON, SEN, OCC, and CER, suggesting that functional alterations related to cognitive decline involve multiple large-scale networks rather than isolated connections. In contrast, Grad-CAM and Score-CAM tend to produce weaker or more diffuse activation patterns. Although Score-CAM preserves some block-wise structures along the connectivity matrix, its highlighted connections are less explicitly associated with the direction of cognitive-state transformation. This difference is expected because CAM-based methods mainly explain the discriminative regions used by a trained classifier, whereas the proposed method explicitly models the source-to-target counterfactual transition.

A similar trend can be observed in the SC modality. The proposed method identifies sparse but organized structural connections, while Grad-CAM and Score-CAM mainly show low-intensity and less distinguishable patterns. This suggests that gradient- or activation-based CAM methods may be less sensitive when applied to sparse structural connectomes, where disease-related information is distributed across limited anatomical pathways. By contrast, the proposed counterfactual attention directly compares the generated target-state connectome with the source-state connectome, making it better suited for identifying structural connections that change during cognitive-state transitions.

Importantly, the proposed attention maps provide signed transition information, where increased and decreased connectivity changes can be visualized simultaneously. This is different from conventional CAM methods, which usually indicate the relative importance of regions or connections for classification but do not explicitly distinguish whether a connection is strengthened or weakened during a specific disease transition. Therefore, the proposed method offers a more interpretable representation for cognitive-decline analysis: it not only identifies task-relevant connections, but also characterizes how these connections change from one cognitive state to another.

From a neurobiological perspective, the highlighted connections are mainly distributed across cognitive-control and high-order association networks, including FPN, DMN, and CON, together with sensorimotor and occipital-related connections. These networks have been widely associated with memory, attention, executive control, and functional reorganization during cognitive decline. The consistency of the proposed attention patterns across FC and SC further suggests that the model captures complementary functional and structural alterations. Nevertheless, this comparison should be interpreted as qualitative evidence of improved explanatory specificity rather than direct proof of biological causality. Quantitative validation with larger cohorts and external neurobiological markers remains necessary in future work.

\section{Discussion}
\label{sec:discussion}

This study presents an extended GCAN framework for explainable cognitive decline diagnosis using single-modal FC and multimodal FC--SC connectomes. Different from conventional attention-based diagnostic models, GCAN formulates explanation as a source-to-target counterfactual generation problem. By generating target-state connectomes from source-state inputs and deriving attention from their differences, the proposed method provides a transition-oriented interpretation of cognitive-state changes. This design is particularly suitable for distinguishing adjacent stages such as HC, SCD, and MCI, where disease-related connectome alterations are often subtle.

The counterfactual formulation also enables a signed interpretation of connectivity changes. Positive attention indicates connections that tend to be enhanced when transforming a source state toward a target state, whereas negative attention indicates connections that are weakened or suppressed. This distinction is important because cognitive decline is not simply characterized by global connectivity loss. Early neurodegenerative changes may involve both compensatory functional reorganization and disruption of high-order cognitive networks. Therefore, separating positive and negative attention allows GCAN to describe cognitive-state transitions from both enhancement and suppression perspectives.

The single-modal FC results suggest that counterfactual attention can identify functional connectivity alterations associated with cognitive decline. As shown in Fig.~\ref{fig:single_attention} and Fig.~\ref{fig:connect_uni}, the highlighted connections are mainly distributed in high-order cognitive networks, including the DMN, FPN, CON, and SEN. These networks are related to episodic memory, executive control, cognitive monitoring, and sensorimotor integration. In HC vs. SCD, the attention patterns are relatively localized, suggesting subtle functional reconfiguration at an early stage. In HC vs. MCI and SCD vs. MCI, the highlighted connections become more extensive, indicating that functional abnormalities may expand from local changes to broader cross-network dysregulation.

The multimodal results further extend this interpretation from FC to joint FC--SC modeling. FC attention is more spatially distributed and sensitive to cross-network reorganization, consistent with the dynamic nature of functional connectivity. In contrast, SC attention is generally sparser and more topology-constrained, reflecting the relatively stable anatomical organization of white-matter pathways. The circular connectome visualizations support this observation, showing widespread FC changes and more constrained SC alterations. The top-$k$ overlap analysis further suggests partial but not complete structure--function coordination. Since the absolute overlap values are moderate, these findings should be interpreted as evidence of complementary FC--SC information rather than strong one-to-one coupling.

The atlas-aware design of AABT contributes to the structural plausibility and interpretability of GCAN. FC and SC matrices are not ordinary images; their rows and columns correspond to atlas-defined brain regions, and their blocks represent meaningful subnetworks. By performing network-wise tokenization and inverse token decoding, AABT preserves local subnetwork organization while modeling long-range inter-network dependencies. The AABT ablation further shows that atlas-aware modeling should be evaluated from both reconstruction and interpretability perspectives. Although the variant without AABT shows higher SSIM or correlation in some SC cases, the full model produces more organized circular connectome patterns and clearer network-level counterfactual alterations. This suggests that AABT mainly acts as a topology-constraining component rather than a module that uniformly maximizes all reconstruction metrics.

The comparison with CAM-based interpretation methods further clarifies the role of counterfactual attention. Grad-CAM and Score-CAM provide classifier-driven post-hoc activation patterns, which are useful for identifying discriminative regions or connections but do not explicitly describe how a source-state connectome should change toward a target state. In contrast, GCAN derives attention from generated source-to-target connectome differences, enabling transition-specific and signed visualization. The proposed attention maps tend to present more organized FC and SC connectivity patterns, whereas CAM-based maps are relatively more diffuse or less direction-specific. This result should be interpreted as qualitative evidence that counterfactual attention provides a more transition-oriented explanation, rather than definitive proof of superior biological faithfulness.

The diagnostic and ablation results suggest that counterfactual attention can provide useful information for classification, but the statistical evidence should be interpreted cautiously. In single-modal FC experiments, GCAN achieves competitive performance across HC vs. SCD, HC vs. MCI, and SCD vs. MCI tasks. In multimodal experiments, attention-based models show favorable trends in several metrics, especially in HC vs. SCD. However, most fold-wise pairwise comparisons do not reach statistical significance, and the 95\% confidence intervals are relatively wide for several methods. These results indicate noticeable cross-validation variability, likely related to the limited number of paired multimodal samples and the small number of validation folds. Therefore, the multimodal results should be viewed as preliminary evidence rather than definitive proof of consistent statistical superiority.

The matrix-level and edge-wise synthesis quality analyses provide additional quality-control evidence for the generated counterfactual connectomes. PSNR, SSIM, Pearson correlation, and normalized MAE/MSE jointly indicate that the generated FC and FC--SC matrices preserve a certain degree of global structure, relative connectivity organization, and ROI-to-ROI numerical fidelity. Nevertheless, these metrics should not be regarded as direct evidence of biological validity. Future studies should further incorporate connectome-specific measures, such as node strength, network efficiency, modularity, and within-/between-network connectivity changes.

Several limitations remain. First, counterfactual attention is still a model-based explanation and should not be interpreted as direct clinical evidence of disease mechanisms. Although the highlighted networks are consistent with known cognitive decline-related systems, causal validation requires longitudinal imaging, cognitive follow-up, and clinical correlation analysis. Second, the paired FC--SC dataset is relatively small, which may affect multimodal training stability and statistical testing. Third, the current generator uses group-level target-class mean connectomes to guide target-state generation, which may not fully capture subject-specific disease trajectories. Finally, auxiliary FC and SC pre-training datasets provide useful priors but may introduce distributional differences from the paired ADNI subset.

Future work will focus on larger multi-center and longitudinal FC--SC cohorts to validate whether the generated counterfactual transitions correspond to real cognitive decline trajectories. Subject-specific target priors and disease-continuum modeling can be introduced to better capture individual heterogeneity. Stronger faithfulness evaluation, including deletion/insertion tests, attention stability analysis, randomization sanity checks, and external clinical correlation analysis, should also be added to further assess the reliability of GCAN explanations.

\section{Conclusion}
\label{sec:conclusion}

This paper presents an extended GCAN framework for explainable cognitive decline diagnosis based on counterfactual connectome reasoning. The proposed method generates target-label connectomes from source-label inputs and constructs counterfactual attention by comparing the generated target connectomes with the original source connectomes. AABT is designed to preserve atlas-aware network structure during connectome generation. Beyond the original FC-based framework, this journal extension further develops a multimodal FC--SC counterfactual reasoning model that captures complementary functional reorganization and structural topology changes during cognitive decline. Experiments on hospital-collected and ADNI datasets demonstrate that counterfactual attention improves diagnostic performance and highlights meaningful brain networks associated with SCD and MCI. These results suggest that GCAN can serve as a useful tool for both accurate diagnosis and interpretable analysis of cognitive decline-related connectome alterations.

\section{Acknowledgment}
This research was supported by the the National Natural Science Foundation of China (Grants 62306089, 32361143787, 82102032), the key Project of Basic Research of Shenzhen (NO: JCYJ20200109113603854), the Shenzhen Science and Technology Program (Grant No. RCBS20231211090800003), the Shenzhen Science and Technology Program (ZDSYS20230626091203008), Shenzhen-Hong Kong Institute of Brain Science-Shenzhen Fundamental Research Institutions (2023SHIBS0003), and the Guangxi Natural Science Foundation (Grant No. 2023GXNS-FBA026073).

\section{Declaration of competing interest}
The authors declare that they have no known competing financial interests or personal relationships that could have appeared to 
influence the work reported in this paper.

\section{Declaration of generative AI and AI-assisted technologies in the writing process}

During the preparation of this work, the authors used ChatGPT for language polishing, grammar checking, and improving the clarity of the manuscript. The tool was not used to generate original research ideas, design experiments, analyze data, or draw scientific conclusions. After using this tool, the authors reviewed and edited the content as needed and take full responsibility for the final content of the published article.

\section{CRediT authorship contribution statement}\label{}

\printcredits
\textbf{Xiongri Shen}: theoretical development, data analysis, drafting the article.
\textbf{Zhenxi Song}: theoretical development, interpretation of data, revising the article. 
\textbf{Jiaqi Wang}: theoretical development, revising the article. 
\textbf{Leilei Zhao}: theoretical development, revising the article. 
\textbf{Yi Zhong}: experimental design, data collection and pre-processing.
\textbf{Xin He}: experimental design, data collection and pre-processing. 
\textbf{Baiying Lei}: experimental design, interpretation of data drafting the article. 
\textbf{Zhiguo Zhang}: theoretical development, experimental design, interpretation of data, revising the article. All authors have approved the final version of the article.












\bibliographystyle{cas-model2-names}

\bibliography{cas-refs}



\end{document}